\RequirePackage[T1]{fontenc}
\documentclass[10pt,journal,compsoc]{IEEEtran}
\usepackage[numbers,sort]{natbib}
\usepackage{amsmath,amsthm,graphicx,booktabs,array,tabularx,microtype,enumitem}
\usepackage{xcolor,url,placeins}
\usepackage[hidelinks]{hyperref}
\newcommand{\carrying}{\textsc{Carrying}}
\newcommand{\matching}{\textsc{Matching}}
\newcommand{\copying}{\textsc{Copying}}

\providecommand{\mathbb}[1]{\mathbf{#1}}
\newsavebox{\bmutablebox}
\newcommand{\fitwidth}[1]{\sbox{\bmutablebox}{#1}\ifdim\wd\bmutablebox>\linewidth\resizebox{\linewidth}{!}{\usebox{\bmutablebox}}\else\usebox{\bmutablebox}\fi}
\newtheorem{proposition}{Proposition}

\hypersetup{pdftitle={The Token Before the Value Is the Key},pdfauthor={Ke Cheng et al.}}
\title{The Token Before the Value Is the Key}
\author{Ke Cheng, Xin Xu, Zhichao Chen, Yixiao Chen, Lei Xin, Jianbo Zhao,\\Fanhu Zeng, Yue Liu, Jun Zhang, and Jie Jiang%
\IEEEcompsocitemizethanks{%
\IEEEcompsocthanksitem K. Cheng, X. Xu, Y. Liu, J. Zhang, and J. Jiang are with AMS, Tencent. K. Cheng is also with CCSE lab, Beihang University. E-mail: \mbox{ckpassenger@buaa.edu.cn}.
\IEEEcompsocthanksitem Z. Chen is with Peking University; Y. Chen is with Data Platform, Tencent. L. Xin, J. Zhao, and F. Zeng are independent researchers.
\IEEEcompsocthanksitem Corresponding author: Jun Zhang (neoxzhang@tencent.com).}}
\IEEEtitleabstractindextext{%
\begin{abstract}
Induction circuits retrieve a value by matching a query to information carried from the token before that value. How this address forms, incorporates context, and supports sequential retrieval remains less clear. We use hybrid language models as a tractable testbed and identify the historical source key as a common causal interface between local preparation and global retrieval. Probes and restoration interventions connect this interface across controlled models, pretrained recurrent and sliding-window hybrids, and released language models. We organize the analysis by address context radius and sequential retrieval depth. First, shared local parameters couple predecessor Carrying to recurrent state processing; separating these demands or biasing initialization promotes address formation. Second, adapting input representations extends an established literal circuit to category matching, making semantically compatible records active competitors. Earlier context resolves this competition through the source key, with a sharp visibility boundary in a controlled single-read model and compositional relays across local layers. Third, fixed relations support memorization shortcuts, whereas sequential generalization requires learning an output-to-query bridge between retrievals. Controlled state-tracking and compositional in-context tasks demonstrate this requirement across stacked stages, looped recurrence, and intermediate-token generation. Together, these results connect circuit formation, contextual disambiguation, and sequential use through a shared retrieval interface.
\end{abstract}
\begin{IEEEkeywords}Induction circuits, mechanistic interpretability, hybrid language models, linear recurrent networks, sliding-window attention, in-context retrieval, sequential computation.\end{IEEEkeywords}}
\begin{document}
\raggedbottom
\maketitle\IEEEdisplaynontitleabstractindextext\IEEEpeerreviewmaketitle
\section{Introduction}
\label{sec:intro}

After reading \emph{Anna Oslo}, a language model can complete a later \emph{Anna} with \emph{Oslo}. The relevant value is stored at the historical Oslo position, but the query refers to the token before it. An induction circuit solves this mismatch by carrying predecessor information into the key at the value position, matching the later query to that key, and copying the associated value~\citep{elhage2021framework,olsson2022induction,singh2024induction}. Thus, ``the token before the value is the key'' describes an address-construction rule: the predecessor supplies the information used to select a historical value, rather than requiring the key vector to equal a token embedding.

\begin{figure}[!t]\centering
\includegraphics[width=\linewidth]{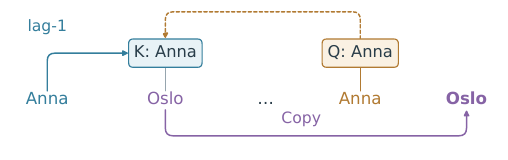}
\caption{\textbf{The predecessor supplies the retrieval address.} In \emph{Anna Oslo}, \carrying\ brings Anna to the historical Oslo position. The later Anna query selects this source through \matching, and \copying\ transmits Oslo. Blue denotes address construction, gold denotes query--key selection, and purple denotes value transmission. The source key $K_{B_1}$ connects local preparation to global retrieval.}
\label{fig:oneih}
\end{figure}

This elementary lookup leaves three questions. How does training build the predecessor-carrying path when other computations can already reduce loss? How does the source address incorporate earlier context when the immediate predecessor is shared by several candidates? And how does one retrieved value become the query for a subsequent lookup? These questions distinguish \emph{forming}, \emph{enriching}, and \emph{composing} retrieval circuits. They also separate having a circuit from learning when and how to use it.

Hybrid architectures provide a useful experimental setting for studying these questions. They interleave recurrent or sliding-window layers~\citep{child2019sparse,beltagy2020longformer,schlag2021linear,yang2025gdn} with global attention~\citep{de2024griffin,lieber2024jamba,ren2025samba}. Their division of computation exposes source preparation and global retrieval at identifiable layer boundaries. We use this structure to observe and intervene on circuits, then compare the same functional operations across full-attention Transformers and released models. Theoretical capacity and architectural complementarity motivate this setting~\citep{wang2025hybrid,cooper2026expressivity,merrill2026olmohybrid}; our focus is how training realizes particular computations within it.

We organize the analysis around a common interface, the historical source key $K_{B_1}$, and two task requirements: \emph{address context radius} $r$ and \emph{sequential retrieval depth} $k$. Radius measures the context assembled into one source address; depth measures causally linked lookups. For example, \emph{Anna chose Oslo} and \emph{Ben chose Rome} require a richer address to distinguish the shared predecessor \emph{chose}. Retrieving \emph{Norway} from \emph{Anna}$\to$\emph{Oslo}$\to$\emph{Norway} instead requires the first result to guide a second lookup.

The paper makes three connected contributions beyond classical $K$-composition~\citep{elhage2021framework,olsson2022induction,singh2024induction} and single-hop hybrid priors~\citep{qiao2026rethinking}:
\begin{enumerate}[leftmargin=*,itemsep=2pt]
\item \textbf{Forming an address.} Cross-architecture interventions identify the source key as the interface through which predecessor Carrying supports retrieval. Controlled branch interventions explain competition with state readout, while local-path design and initialization influence emergence and stage placement (Sections~\ref{sec:interface}--\ref{sec:formation}).
\item \textbf{Enriching an address.} Representation adaptation extends literal matching to categories and introduces competition between compatible sources. Earlier context resolves that competition through the source key; controlled visibility tests and adjacent relays connect this ability to the address radius (Section~\ref{sec:context}).
\item \textbf{Composing retrievals.} Corpus audits and controlled tasks distinguish reusable sequential computation from memorization and single-hop shortcuts. Training the output-to-query bridge enables sequential generalization across stacked stages, recurrence, and intermediate-token generation (Section~\ref{sec:sequential}).
\end{enumerate}

\section{A Common Interface for Retrieval Circuits}
\label{sec:interface}\label{sec:method}

\subsection{Functional Roles and Experimental Settings}
\label{sec:allocation-probes}
We distinguish three functions. \carrying\ supplies predecessor or contextual information to a historical source position. \matching\ uses the query to select a source key. \copying\ transmits the selected value. A \emph{retrieval stage} comprises a local/recurrent preparation block followed by its immediately succeeding global-attention receiver (e.g., $\mathrm{Stage\ 1}=\mathrm{L0}\to\mathrm{L1}$ and $\mathrm{Stage\ 2}=\mathrm{L2}\to\mathrm{L3}$ in 4-layer controlled hybrids; $\mathrm{Stage\ 1}=\mathrm{L0\text{--}L2}\to\mathrm{L3}$ and $\mathrm{Stage\ 2}=\mathrm{L4\text{--}L6}\to\mathrm{L7}$ in 8-layer $3{:}1$ pretrained hybrids); a stage can be present in the architecture without carrying out all of these functions.

We compare \emph{recurrent hybrids} based on DeltaNet / Gated DeltaNet (GDN)~\citep{schlag2021linear,yang2024gla,yang2025gdn}, \emph{sliding-window hybrids} based on sliding-window attention (SWA)~\citep{child2019sparse,beltagy2020longformer}, and \emph{full-attention Transformers}~\citep{vaswani2017attention}. Small controlled memory--attention models ($d{=}128$, 4 heads) isolate branch gradients and interventions; pretrained 79M (8 layers, $d{=}512$) and 318M (16 layers, $d{=}1024$) models trained for 30K steps on packed 1,024-token OpenWebText sequences test the same probes across parameter scales; released Qwen models provide an additional architectural comparison~\citep{qwen3model,qwen35model}. Appendix~\ref{app:settings} lists full configurations and replication, and Appendix~\ref{app:names} maps descriptive task names to the original evaluation keys.

\begin{figure*}[!t]\centering
\includegraphics[width=\textwidth]{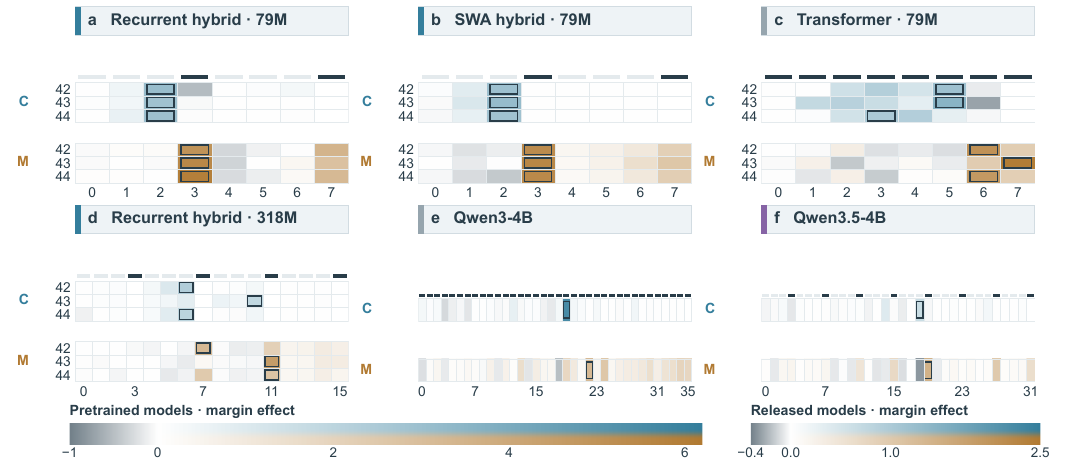}
\caption{\textbf{Source preparation and retrieval across architectures and scales.} Top: matched pretrained architectures. Bottom: the larger recurrent hybrid and released Qwen models. Blue denotes Carrying and gold Matching. Pretrained rows retain all three training seeds; released models retain their layer coordinates. Columns index layers. Outlines mark row maxima; dark strips mark global attention. Each color scale is shared only within its labeled model group.}
\label{fig:distribution}\label{fig:archcontrol}\label{fig:v33-framework}
\end{figure*}

The layer maps expose a consistent source--receiver organization in hybrids (Fig.~\ref{fig:distribution}). \carrying\ concentrates in efficient layers preceding a global receiver, while \matching\ and \copying\ concentrate within that receiver. Head-level interventions further align the matching and copying functions (Appendix Fig.~\ref{fig:headcontrol}). These observations make hybrid blocks a tractable setting for studying how the functions connect.

\subsection{Causal Probes and the Historical Source Key}
Different layer types do not share comparable internal attention maps. We instead intervene on the residual block update $u_\ell(p)=h_\ell(p)-h_{\ell-1}(p)$ at layer $\ell$ and position $p$~\citep{geiger2023causal,zhang2024patching}. In a clean prompt $c=AB\ldots CD\ldots A_q$, the target value $B$ occurs at source position $s=B_1$, and $D$ is the distractor. We measure the logit margin $m=z_B-z_D$ and replace a clean block update with an RMS-matched donor update:
\begin{equation}
 \widetilde u_d=u_d\frac{\operatorname{RMS}(u_c)}{\operatorname{RMS}(u_d)+10^{-8}},
 \qquad \delta=m_{\rm clean}-m_{\rm patched}.
 \label{eq:patch}
\end{equation}
The Carrying probe changes the immediate predecessor ($\tau{=}1$) and patches the historical value position $s{=}B_1$. A lag-$\tau$ probe changes the token $\tau$ positions before the value and patches that same source position. Controlled stage-level Carrying is measured at the local layer immediately before its receiver. Pretraining summaries instead use block Carrying, $C_{\mathcal B}=\sum_{\ell\in\mathcal B}\max(C_\ell,0)$, over the local layers of each probed block. The probed blocks are identified by their receiver layers: L3/L7 at 79M and L7/L11 at 318M (Appendix~\ref{app:probe-definitions}). The Matching probe swaps historical associations ($AB\ldots CD\mapsto CB\ldots AD$) and patches the query update at $q$. For head-level \copying\ (Appendix Fig.~\ref{fig:headcontrol}), the donor replaces the historical value tokens ($B,D\mapsto B',D'$) while holding keys, query, and attention weights fixed, and patches the value-readout contribution at $q$. Unscaled activation patching without RMS normalization preserves the same peak layer and head coordinates across tested checkpoints (Appendix~\ref{app:protocols}). Appendix~\ref{app:probe-definitions} summarizes the probes; Table~\ref{tab:sourcekey} reports source-key restoration.

\begin{table*}[!t]\centering\small\setlength{\tabcolsep}{6pt}
\caption{\textbf{Source-key restoration recovers retrieval across model settings.} Loss and gain are measured on the same corrupted background. Controlled and 79M entries report training-seed mean $\pm$ SD; Qwen3.5-4B gives paired-prompt means with bootstrap intervals; the larger recurrent hybrid gives point estimates. Specificity controls are reported in Appendix Table~\ref{tab:sourcekey-specificity}.}
\label{tab:probe-operations}\label{tab:sourcekey}
\begin{tabularx}{\textwidth}{@{}Xlrrr@{}}\toprule
Model & Interface & Full-path loss & Source-key gain & Recovery (\%)\\\midrule
\multicolumn{5}{@{}l}{\textit{Margin change}}\\
Controlled pair & L0$\to$L1 & $4.42\pm0.81$ & $\mathbf{+4.42\pm0.80}$ & 100.0\\
Controlled $3{:}1$ block & L2$\to$L3 & $6.49\pm1.02$ & $\mathbf{+6.37\pm0.98}$ & 98.2\\
Recurrent hybrid (79M) & L2$\to$L3 & $2.71\pm0.08$ & $\mathbf{+2.68\pm0.09}$ & 98.9\\
Recurrent hybrid (318M) & L7, L11$^\dagger$ & $2.22$ & $\mathbf{+1.84}$ & 83.1\\
Qwen3.5-4B & L18$\to$L19 & $1.44\ [1.30,1.59]$ & $\mathbf{+1.42}\ [1.28,1.57]$ & 98.3\\
\midrule
\multicolumn{5}{@{}l}{\textit{NLL change ($\Delta\mathrm{NLL}$)}}\\
SWA hybrid (79M, $W{=}2$) & L2$\to$L3 & $1.29\pm0.07$ & $\mathbf{+1.30\pm0.07}$ & 100.2\\
SWA hybrid (79M, $W{=}8$) & L2$\to$L3 & $1.33\pm0.05$ & $\mathbf{+1.33\pm0.06}$ & 100.5\\
\bottomrule
\end{tabularx}
\par\smallskip\parbox{\textwidth}{\footnotesize $^\dagger$The larger-model margin entry sums receiver-specific gains after a shared upstream disruption. Its recovery is a descriptive aggregate. The separate single-receiver NLL test and all control details appear in Appendix Table~\ref{tab:sourcekey-specificity}.}
\end{table*}

To test how upstream preparation reaches the receiver across hybrid families, we disrupt Carrying at the source window and substitute cached clean activations inside the global attention receiver, using the source--receiver boundaries identified by the localization analysis in Fig.~\ref{fig:distribution}. Single-receiver recovery is the ratio of mean restoration gain to mean disruption loss. In the larger recurrent hybrid, Table~\ref{tab:sourcekey} separately labels the sum of two receiver-specific gains. Across settings, source-key restoration produces substantially larger recovery than the source-value, query, and off-site-key controls (Appendix Table~\ref{tab:sourcekey-specificity}). These interventions identify the historical source key as a causal interface between local preparation and global retrieval.

\subsection{Address Radius and Retrieval Depth}
\label{sec:coords-theory}
Let $S$ denote the causal one-position shift, $(SX)_t=X_{t-1}$. A source address can combine several shifted features, while a global lookup selects among historical keys:
\begin{align}
 K_j&=A_r(X)_j=\phi_r\big(X_j,SX_j,\ldots,S^rX_j\big),\label{eq:address}\\
 \mathcal R(Q_q;K,V)&=\sum_{j<q}\alpha(Q_q,K_j)V_j.\label{eq:retrieval}
\end{align}
The \emph{address context radius} $r$ describes the preceding context available to a source key. The \emph{sequential retrieval depth} $k$ counts causally chained lookups:
\begin{equation}
 y_1=\mathcal R_1(q_1),\qquad y_2=\mathcal R_2(y_1),
 \label{eq:serial}
\end{equation}
where the retrieved representation from the first stage supplies the query for the second stage through the residual stream. Adding parallel readouts from the initial query $q_1$ does not create this sequential dependency.

The stylized analytical models in Appendix~\ref{app:theory} formalize three idealizations that guide our empirical diagnostics: Proposition~\ref{prop:locality} bounds local address radius in a stateless local subnetwork before the first global receiver and describes composition of adjacent shifts; Proposition~\ref{prop:demand} relates incremental risk reduction in nested linear feature spaces to gated path gradients; and Proposition~\ref{prop:memorization} analyzes how an explicit instance-indexed lookup absorbs training residuals and attenuates gradients to reusable features.

\section{Forming the Source Address}
\label{sec:formation}

The first requirement is to make predecessor information available at the historical value position. A recurrent layer can also reduce retrieval loss through its own state readout. We ask how these routes interact during training and how local-path design changes the formation of Carrying.

\subsection{Predecessor Access Changes Formation Time}
\begin{figure*}[!t]\centering
\includegraphics[width=\textwidth]{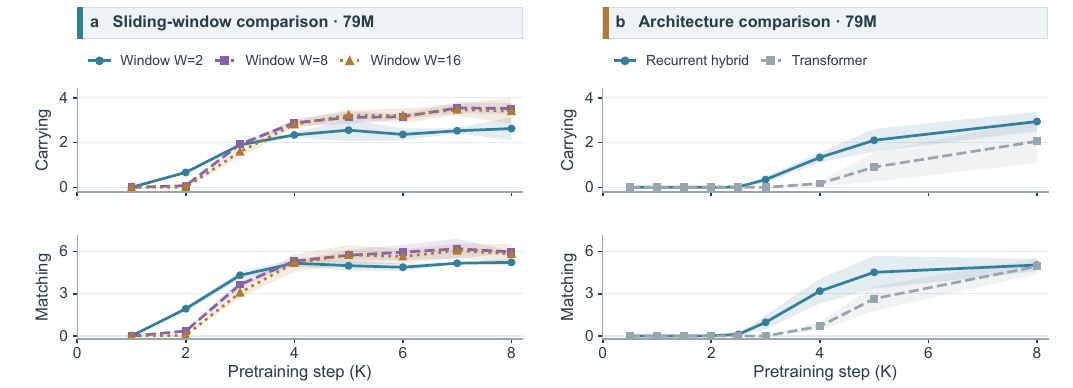}
\caption{\textbf{Access to the predecessor changes circuit acquisition.} Window variants and the natural-text architecture comparison are shown separately. Carrying and Matching are margin effects at fixed endpoint-selected layers throughout training. Curves show three-seed means with sample-SD bands at measured checkpoints.}
\label{fig:early-formation}
\end{figure*}

At a matched pretraining scale, we compare sliding-window variants and separately compare recurrent hybrids with full-attention Transformers under natural-text training. Carrying and Matching are followed at each run's fixed endpoint-selected layers (Fig.~\ref{fig:early-formation}). The narrowest window advances predecessor Carrying relative to wider windows; recurrent hybrids develop both operations earlier than the full-attention control. Matching onset can tie the intermediate window at the weakest thresholds. Appendix~\ref{app:formation-timing} defines the checkpoint-based comparison.

\subsection{Recall Can Shift Between Computational Routes}
\begin{figure*}[!t]\centering
\includegraphics[width=\textwidth]{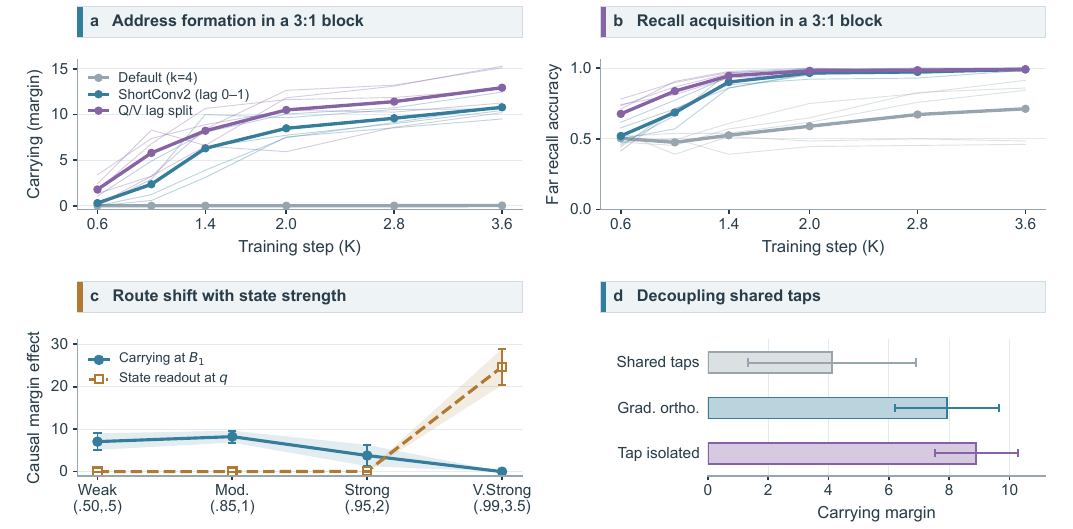}
\caption{\textbf{Local-path interventions separate circuit formation from alternative recall routes.} a,b: In a controlled three-layer preparation block, thin curves retain individual seeds and thick curves show condition means. c: A separate memory--attention pair shifts from predecessor Carrying to recurrent state readout as state strength increases. d: Direct gradient orthogonalization and tap isolation increase Carrying in the strong-state pair. Bars show means with sample SD. Full numerical settings and controls appear in Appendix Table~\ref{tab:state}.}
\label{fig:formation}\label{fig:v32-circuit12}
\end{figure*}

In a controlled memory--attention model, we vary state retention and readout gain while keeping the task mixture and shared short convolution fixed. A standalone recurrent stack provides the corresponding comparison without a global receiver. We then separately patch the historical source update and the recurrent update at the query, distinguishing source-address preparation from direct state readout (Fig.~\ref{fig:formation}c; Appendix Table~\ref{tab:state}).

Stronger state retention improves standalone recurrent recall, but its effect in a hybrid depends on which route the model uses. In the strongest-state regime, high recall coexists with weak predecessor Carrying and a dominant recurrent query-readout effect. The intermediate setting $(\gamma,g)=(.95,2.0)$ exhibits elevated between-seed variance consistent with a transitional regime in which only a subset of seeds complete cross-layer emergence within 1,400 steps while the remainder begin shifting toward intra-layer state readout. Thus, behavioral recall alone does not identify whether a cross-layer induction circuit has formed.

\subsection{Shared Taps Couple State Processing and Carrying}
To locate the interaction in the controlled memory block, we decompose the total parameter gradient $g_{\mathrm{total}}=g_{\mathrm{loc}}+g_{\mathrm{state}}$ on the shared short convolution into its local residual branch $g_{\mathrm{loc}}$ (which supplies predecessor features at $B_1$ for cross-layer Carrying) and its recurrent state branch $g_{\mathrm{state}}$ (which writes and reads the intra-layer state $M_t$) across tasks at the same checkpoints. For shared-parameter gradients $g_0,g_1$ weighted by $\pi_0,\pi_1$, a joint step changes the Carrying objective by
\begin{equation}
 \Delta\mathcal L_1=-\eta\big(\pi_1\|g_1\|^2+\pi_0g_1^\top g_0\big)+O(\eta^2).
 \label{eq:conflict}
\end{equation}
While checkpoint-averaged lag-one branch cosines remain nonnegative, several taps exhibit negative branch inner products, and position-decomposed directional derivatives on the relative lag-one-to-lag-zero weight ratio during the pre-formation window show that non-source state readouts pull shared taps toward \texttt{lag-0} across all seeds (Appendix Table~\ref{tab:gradients}). In the coupled residual-sharing surrogate of Appendix~\ref{app:gates}, an alternative state readout $u$ that absorbs a fraction of the target residual also raises the escape threshold required for a multi-gate cross-layer path to leave the origin.

To test whether shared-parameter coupling directly impedes Carrying in this controlled strong-state regime, we perform two direct decoupling interventions (Appendix Table~\ref{tab:gradients}D): (i) \emph{gradient orthogonalization}, which projects $g_{\mathrm{state}}$ onto the orthogonal complement of $g_{\mathrm{loc}}$ whenever $g_{\mathrm{loc}}^\top g_{\mathrm{state}}<0$ on shared convolution taps; and (ii) \emph{head/tap parameter isolation}, which routes state writing through a dedicated \texttt{lag-0} projection while isolating predecessor taps for cross-layer Carrying at matched initial activation norm. Both interventions substantially increase continuous Carrying magnitude and recall accuracy across seeds, consistent with shared-tap interference acting as a bottleneck on cross-layer address formation in the controlled setting.

\subsection{Query-Path History in Pretrained Hybrids}
We next test whether pretrained recurrent hybrids rely equally on query and value short-convolution history at the source window. DeltaNet / Gated DeltaNet blocks pass short-convolved queries, keys, and values into a delta-rule recurrent state update~\citep{schlag2021linear,yang2024gla,yang2025gdn}. Unlike the controlled additive memory block, pretrained recurrent-hybrid checkpoints already develop baseline source Carrying under default initialization; we therefore apply inference-time source-window lesions that selectively remove historical taps from the query or value projection while leaving the rest of the checkpoint intact (Table~\ref{tab:s1c-qkv}).

Across both pretraining scales, removing value-path history at the source window largely preserves unambiguous recall and contextual disambiguation, whereas removing query-path history strongly disrupts both. Retaining only the current and immediately preceding offsets also preserves most of the disambiguating margin. These lesion results show that, at the evaluated checkpoints, source-address construction depends primarily on historical taps along the query pathway rather than the value pathway, consistent with using current-token value features and history-dependent queries when reading local context into $K_{B_1}$.

\begin{table*}[!t]\centering\small\setlength{\tabcolsep}{7pt}
\caption{\textbf{Query-path history supports source-address construction.} Lesions remove specified convolution taps throughout the sequence or within the source window. The clean row reports margins; other rows report intervened-minus-clean margin. Paired prompts are evaluated at one checkpoint per scale.}
\label{tab:s1c-qkv}
\begin{tabular}{@{}lrrrr@{}}\toprule
 & \multicolumn{2}{c}{Scale: 79M} & \multicolumn{2}{c}{Scale: 318M}\\
Intervention & Recall & Disambig. & Recall & Disambig.\\\midrule
Clean margin & 2.54 & 3.12 & 2.22 & 3.42\\\midrule
All positions: remove history & $-2.81$ & $-3.37$ & $-2.30$ & $-3.38$\\
Source window: remove Q/K/V history & $-2.67$ & $-3.31$ & $-2.19$ & $-3.32$\\
Source window: retain current + predecessor & $+0.23$ & $-0.43$ & $-0.16$ & $+0.04$\\
Source window: remove V history & $-0.34$ & $+0.11$ & $+0.22$ & $-0.17$\\
Source window: remove Q history & $-2.69$ & $-3.46$ & $-2.04$ & $-3.31$\\\bottomrule
\end{tabular}
\end{table*}

\subsection{Directional Initialization in Controlled Models}
\label{sec:softprior-cross-scale}\label{sec:scale}
Motivated by the complementary roles of current-token and predecessor taps, \emph{SoftPrior initialization} biases local taps toward current-token state processing on a subset of heads and predecessor Carrying on the remaining heads while keeping all taps trainable. Unlike hard tap masks, the initial bias can be revised during optimization. Appendix~\ref{app:softprior} gives the initialization coefficients and norm-controlled ablations.

Norm-controlled ablations separate directional lag preference from filter scale (Appendix Table~\ref{tab:initialization-acquisition}). Across matched seeds, both standard SoftPrior and a norm-preserved variant improve Carrying and composite recall margins over default, isotropic norm-matched, and swapped-lag controls. Directional predecessor bias therefore promotes address formation even at the default filter norm; changing the norm contributes an additional effect.

Pretraining comparisons across three independent seeds per scale (42, 43, and 44 at both 79M and 318M) are reported in Appendix~\ref{app:pretraining-readouts} and Appendix Table~\ref{tab:pretraining-full}: SoftPrior does not produce a statistically resolvable net change in aggregate perplexity ($+1.65$ at 79M, $t{=}1.85$; $-1.95$ at 318M, $t{=}1.01$), but increases earlier-receiver Matching in all $6/6$ seeds across both scales and induces a $3/3$-seed near-for-far trade-off at 79M (trading near-range recall for improved distant recall, distant disambiguation, and code-identifier reuse).

At the evaluated early-training budget, short-window hybrids acquire retrieval more readily than the configured full-attention control. Selective early attenuation of predecessor taps can move the dominant circuit to a later stage (Appendix~\ref{app:swa-placement}). Controlling the token-before-value path thus changes not only whether retrieval forms, but where it is implemented.

\section{Enriching the Source Address}
\label{sec:context}\label{sec:recall}

In \emph{Anna chose Oslo} and \emph{Ben chose Rome}, the immediate predecessor \emph{chose} is shared. A query that repeats \emph{Anna chose}---or a category-level query whose immediate cue matches several same-category records---must distinguish the historical sources using earlier context (Fig.~\ref{fig:context}). This is a requirement for a larger address radius ($r\ge 2, k{=}1$), not an additional retrieval step. We trace three connected questions along the $(r, k{=}1)$ axis: (i) whether an established literal induction circuit can broaden its matching from exact word identity to category compatibility solely through input representations, and how that broader matching induces source-key competition among category-compatible but answer-conflicting records; (ii) how earlier identity cues at lags $d\in\{2,3,4\}$ enter $K_{B_1}$ to resolve that competition, both in a controlled single-read writer and through composed adjacent relays; and (iii) where local context propagation stops in stateless sliding-window stacks versus recurrent hybrids.

\subsection{Semantic Matching and Candidate Competition}
\label{sec:semantic-competition-disambig}
We first adapt an established literal induction circuit by updating only cue-token embeddings, leaving the attention, MLP, position, and output weights frozen. On readable adjective--label prompts, an unseen-in-prompt query word such as \emph{pleasant} retrieves the label associated with a same-category word such as \emph{wonderful}. The adapted model acquires category retrieval while retaining its dependence on the predecessor-carrying edge. Patching adapted query and source-key activations together reproduces this behavior in the initial model; either activation patch alone gives only partial transfer, and source-value replacement leaves performance essentially unchanged (Table~\ref{tab:semantic-competition-disambig}). Input representations can therefore broaden matching while reusing an established Carrying--Matching--Copying circuit. This adaptation also reduces exact-match selectivity on literal prompts (Appendix~\ref{app:semantic-literal}), motivating the competition test below.

\begin{figure*}[!t]\centering
\includegraphics[width=\textwidth]{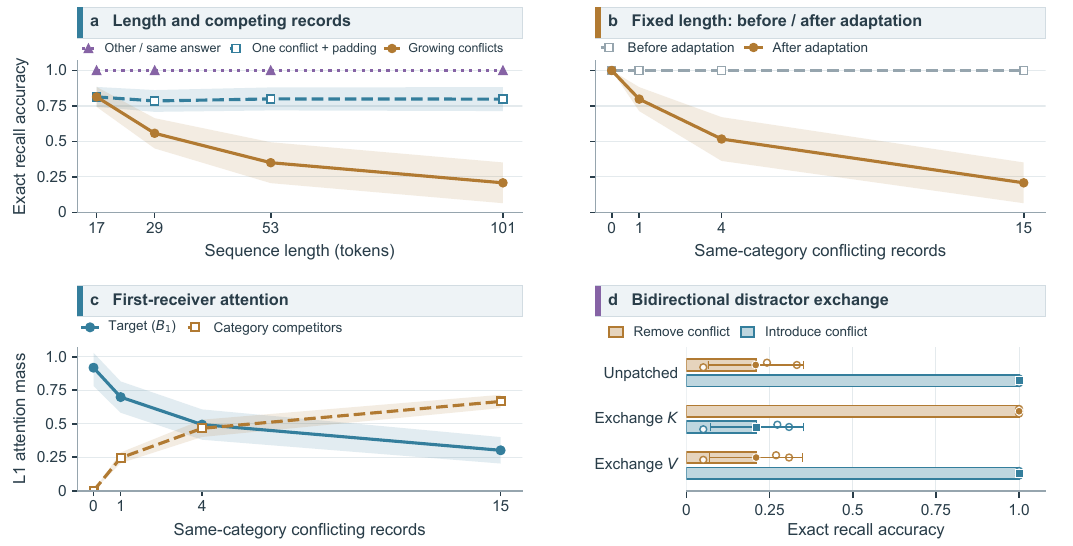}
\caption{\textbf{Compatible source keys create retrieval competition.} a: Context grows with either one conflict or a growing set of conflicts. b: At fixed context length, category adaptation changes sensitivity to competing records. c: Target and competitor attention at the first receiver track source selection in the same fixed-length experiment. d: Paired key exchanges remove conflict from conflicting inputs and introduce it into cross-category inputs; value exchanges are controls. Curves and bars show means; bands and whiskers show sample SD.}
\label{fig:semantic-competition}
\end{figure*}

Broader matching also creates competitors. Category adaptation assigns a common answer to same-category cues within each training prompt; the competition test assigns conflicting answers while retaining the exact query-word source as the target. Adding same-category records with conflicting answers reduces exact recall, while cross-category distractors and same-category records sharing the target answer preserve it. Holding the conflicting-record count fixed keeps performance stable as cross-category padding lengthens the prompt (Fig.~\ref{fig:semantic-competition}a,b). Target and competitor attention at the first receiver track the redistribution of source selection (Fig.~\ref{fig:semantic-competition}c). Bidirectional interventions locate this effect at distractor source keys: replacing conflicting keys with paired cross-category keys restores recall, and transplanting conflicting keys into otherwise clean inputs induces the failure. Corresponding value replacements do not reproduce either effect (Fig.~\ref{fig:semantic-competition}d). Thus, competitor compatibility explains the decline more directly than length alone.

\subsection{Earlier Context Resolves Candidate Competition}
We next isolate disambiguation in a controlled single-read model: one local source writer, one global reader restricted to historical code positions, and two positionwise blocks. Two records share either an identical cue word or different category-related words, while an earlier person name identifies the target. The query contains the target identity and category. This configuration exposes the context available to each source key without a second global retrieval route.

\begin{figure*}[!t]\centering
\includegraphics[width=\textwidth]{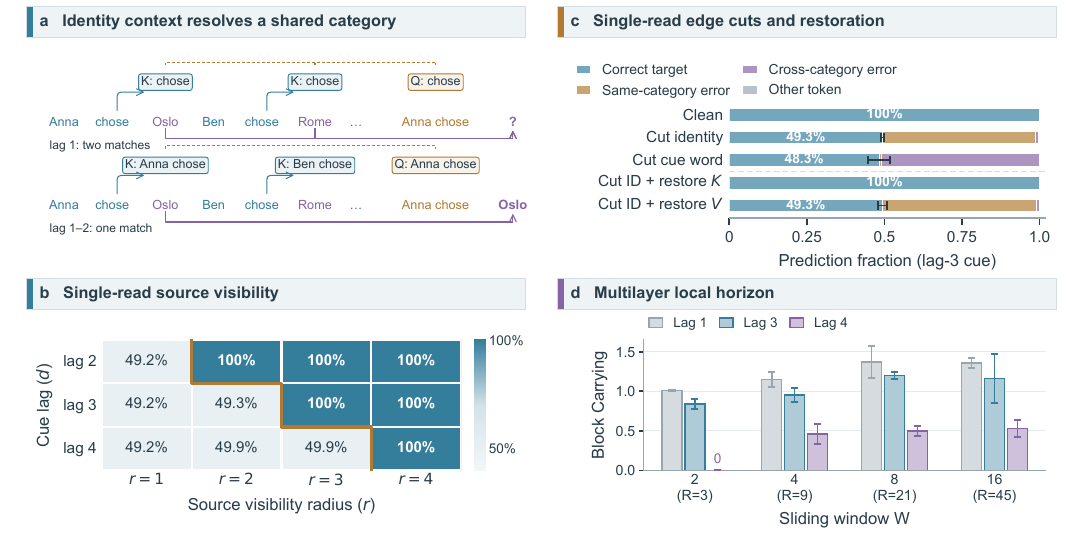}
\caption{\textbf{Context resolves retrieval competition within the accessible local horizon.} a: The shared predecessor supplies category compatibility; earlier identity context selects the source. b,c: In the controlled single-read model, visibility bounds identity selection, and clean-key restoration reverses identity-edge cuts. Stacks show mean prediction fractions with sample-SD accuracy whiskers; K/V restoration follows an identity-edge cut. d: In pretrained SWA hybrids, lag-three and lag-four effects are measured within the pre-receiver local block. The narrowest window reaches lag three but excludes lag four. Bars show means with sample SD. Full visibility, edge-cut, and horizon controls appear in Appendix Table~\ref{tab:context-access}C--F.}
\label{fig:context-disambig}\label{fig:context}\label{fig:v32-circuit3}
\end{figure*}

If the identity cue lies beyond the writer's radius, swapping the two identities preserves all reader keys, values, and the query while changing the correct answer. Proposition~\ref{prop:indistinguishability} formalizes the resulting paired-accuracy bound. The visibility scan follows this boundary for both identical and category-related words (Fig.~\ref{fig:context-disambig}b; Appendix Table~\ref{tab:context-access}). Under full visibility, cutting the identity-reading edge disrupts within-category selection; cutting the immediate word edge instead disrupts category selection. Restoring clean source keys after an identity-edge cut recovers retrieval, whereas restoring values does not (Fig.~\ref{fig:context-disambig}c). Identity-edge cuts mainly select a wrong source within the target category, whereas word-edge cuts mainly select another category. The predecessor supplies category compatibility, and earlier context distinguishes compatible sources. Appendix~\ref{app:semantic-details} reports the full visibility and intervention results.

\begin{table}[!t]\centering\small\setlength{\tabcolsep}{4pt}
\caption{\textbf{Embedding adaptation extends literal retrieval to categories.} Only cue embeddings are trained. Q/K activation patches enter the initial model; edge-content interventions enter the adapted model. Accuracy is mean $\pm$ training-seed SD (\%).}
\label{tab:semantic-competition-disambig}
\begin{tabularx}{\linewidth}{@{}Xr@{}}\toprule
Condition & Category accuracy\\\midrule
Initial literal circuit & $36.1\pm5.0$\\
Adapted cue embeddings & $100.0\pm0.0$\\
\quad Replace predecessor-edge content & $46.9\pm3.0$\\
\quad Replace unrelated-edge content & $100.0\pm0.0$\\\midrule
Transfer adapted source keys & $50.6\pm1.5$\\
Transfer adapted query & $62.0\pm5.3$\\
Transfer both query and keys & $100.0\pm0.0$\\
Transfer adapted source values & $36.1\pm5.0$\\\bottomrule
\end{tabularx}
\end{table}

\subsection{Candidate Selection and Adjacent Relays}
\label{sec:relay-mechanism}
While the single-read disambiguation experiment uses a single-layer local writer with direct access to $\mathrm{lag\text{-}1..4}$, multi-layer hybrids can also assemble longer addresses by composing short local shifts across successive layers. On repeated-anchor prompts, we separate attention assigned to plausible sources from selection within that set: candidate mass is $P_{\rm cand}=P(B_1)+P(B_2)$, while conditional target selection is $R_{\rm in}=P(B_1)/P_{\rm cand}$. In the controlled hybrid, swapping only the earlier context $(c_1,c_2)\leftrightarrow(c_3,c_4)$ preserves candidate coverage $P_{\rm cand}$ while reversing within-candidate preference $R_{\rm in}$ and flipping the logit margin; restoring the clean target source key $K_{B_1}$ restores the preference toward the target candidate. Tracking $P_{\rm cand}$ and $R_{\rm in}$ over training verifies that candidate localization precedes contextual selection: during early training, predecessor Carrying rapidly concentrates attention onto $P_{\rm cand}$ while $R_{\rm in}$ remains at chance; only after candidate mass saturates does $R_{\rm in}$ separate the target from the distractor, and suppressing $P_{\rm cand}$ via symmetric candidate logit biases stalls subsequent lag-2 tap growth relative to an off-candidate control.

Two interventions identify this multi-layer relay path. First, removing longer short-convolution taps while preserving current-token and predecessor taps largely preserves contextual disambiguation in pretrained recurrent hybrids (Table~\ref{tab:s1c-qkv}). Second, a same-background factorial intervention crosses intermediate-anchor restoration with blocking the next adjacent transfer: restoring the intermediate representation recovers retrieval when the downstream relay remains open, but gives little recovery when that relay is blocked (Appendix Table~\ref{tab:relay}E). With $S$ denoting one-position shift, successive relays compose as $S\circ S=S^2$ (Proposition~\ref{prop:locality}).

\subsection{Local Horizon and Recurrence}
For a stateless local subnetwork of $L$ layers with window $W$~\citep{child2019sparse,beltagy2020longformer} preceding the \emph{first} global receiver (where no global attention or cross-position recurrent state has yet mixed distant positions), the source representation at $B_1$ depends only on
\begin{equation}
 R_{\rm local}\le L(W-1).
 \label{eq:radius}
\end{equation}
Proposition~\ref{prop:locality} follows this dependency through successive local layers. In an 8-layer $3{:}1$ SWA hybrid ($[\mathrm{SWA}^3,\mathrm{FullAttn}]^2$), the three local layers $\mathrm{L0\text{--}L2}$ before the first global receiver $\mathrm{L3}$ form a strictly local subnetwork with $L{=}3$. For $W{=}2$, Eq.~\eqref{eq:radius} excludes lag-4 information from this block. A positive lag-3 effect separately tests whether training uses context within the reachable radius.

Figure~\ref{fig:context-disambig}d and Appendix Table~\ref{tab:swa}F test both sides of this boundary across the 79M SWA window family $W\in\{2,4,8,16\}$ by reporting peak lag-1 Carrying ($C_1$), in-horizon lag-3 Carrying ($\Sigma_3^{\mathrm{blk1}}$), and out-of-horizon lag-4 Carrying ($\Sigma_4^{\mathrm{blk1}}$) summed strictly within the pre-receiver local block $\mathrm{L0\text{--}L2}$, alongside all-layer lag-4 Carrying ($\Sigma_4^{\mathrm{all}}\ge\Sigma_4^{\mathrm{blk1}}$). For $W{=}2$, in-horizon lag-3 Carrying remains active ($\Sigma_3^{\mathrm{blk1}}>0$) whereas out-of-horizon lag-4 Carrying in $\mathrm{L0\text{--}L2}$ drops identically to zero, while wider windows $W\in\{4,8,16\}$ (for which $R_{\mathrm{local}}\ge 9$) retain positive lag-4 Carrying and recurrent hybrids exhibit longer-range lag effects, consistent with context propagation through their local layers and persistent state $M_t$ (Appendix Fig.~\ref{fig:relay-full}c).

\section{Composing Retrieval Stages}
\label{sec:sequential}

Address enrichment increases the context assembled into a single source key ($r\ge 2, k{=}1$). We now turn to sequential retrieval depth ($k{=}2$), separating the availability of multiple induction stages from learning to pass one retrieved value into the next query. Because relational depth in a task specification need not coincide with the contextual retrieval depth actually used by a model---a learned parametric association, a surface inconsistency heuristic, or local span copying can supply one relation without an additional contextual lookup---we combine our stylized optimization models (Propositions~\ref{prop:demand}--\ref{prop:memorization}) with natural-corpus ladder audits, value-consistency shortcut controls, cross-layer relay patching, and controlled state-tracking tasks inspired by StateBench's hierarchy~\citep{petty2026statebench} and four compositional in-context learning (ICL) tasks.

\subsection{Task Demand and Memorization Shortcuts}
\label{sec:starvation-memorization}
In our stylized feature-space model (Proposition~\ref{prop:memorization}, Part~1), parallel modules restricted to single-hop feature spaces $\mathcal H_1$ leave an unresolved risk gap $\Delta_2(D)>0$ on tasks requiring higher-order features $\mathcal H_2$. Proposition~\ref{prop:demand} further shows that along a gated multi-interface path $a(\theta)=\prod_e\theta_e$, the gradient driving a new stage scales with the residual task demand $\Delta_k$ left unresolved by earlier computations:
\begin{equation}
 \frac{\partial\mathcal L}{\partial\theta_e}
 =\left(a-\sqrt{2\Delta_k}\right)\prod_{e'\ne e}\theta_{e'}.
 \label{eq:gates}
\end{equation}
A literal chain in a corpus need not require sequential retrieval. We therefore audit candidate relations in OpenWebText and Python code, progressively removing shortcuts such as direct copying, repeated spans, and stable attribute priors (Appendix Table~\ref{tab:corpus-ladder}). In the sampled text windows, the cumulative filter leaves no shortcut-free candidates; manual inspection of the alternative filter's survivors identifies repeated phrases or lists. The code audit similarly exposes direct-copy and API-prior routes. These audits characterize the sampled candidate constructions, motivating training data that explicitly separates sequential retrieval from its shortcuts.

\begin{figure*}[!t]\centering
\includegraphics[width=\textwidth]{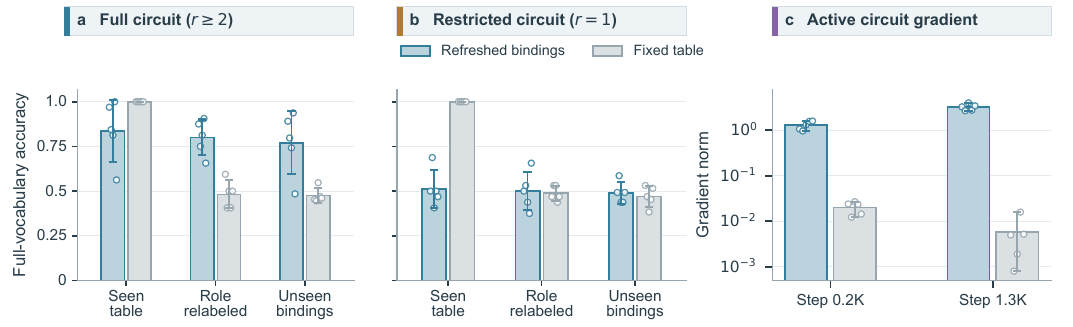}
\caption{\textbf{Fixed relational tables favor memorization over reusable retrieval.} a,b: Generalization under full and restricted structures, comparing dynamically refreshed relational bindings with a repeated $N{=}32$ table. Bars show means and sample SD. Role-preserving token relabeling $\pi$ and unseen instances test whether the model learned a reusable circuit or memorized token bindings. c: Active circuit gradients in the full structure at matched early and late checkpoints (log scale). Bars show means; gradient whiskers span the observed seed range.}
\label{fig:v32-circuit6}
\end{figure*}

Controlled memorization experiments test the corresponding optimization effect. Repeating a fixed relational table allows both full and restricted structures to fit training instances, but performance drops under role-preserving relabeling and unseen bindings. Gradients to the reusable pathway also diminish (Fig.~\ref{fig:v32-circuit6}), matching the residual-absorption mechanism in Proposition~\ref{prop:memorization}. Refreshing bindings maintains pressure on that pathway.

Stage availability is a separate issue. Controlled serial training concentrates Carrying in the first stage, while stagewise-to-serial training preserves a later stage that remains useful when the first is ablated (Fig.~\ref{fig:twoih}b; Appendix Table~\ref{tab:recruit}). These controlled interventions distinguish the recruitment of multiple stages from their subsequent composition.

\subsection{Shortcuts Versus Sequential Generalization}
\label{sec:twohop-bridge}
We compare training routes on two contextual chains queried at their initial node (Table~\ref{tab:twohop}). A single stage can memorize a fixed table. It can also combine one contextual lookup with a stable parametric mapping $\psi$, or exploit an inconsistency heuristic when only the target relation is refreshed. A value-consistency control grid and separate record ablations distinguish these solutions: the parametric route ignores the second record, while the inconsistency route ignores the first. Both fail when the evaluation changes the relation that their shortcut assumes.

Neither stage availability nor additional endpoint training automatically establishes composition. Two stages trained only on single-hop tasks, a model trained from scratch on endpoints, and parallel readouts of the initial query all fail to generalize the two-hop relation. In contrast, refreshed supervision of both the intermediate and endpoint supports unseen-chain generalization through the ordinary residual stream.

\begin{figure*}[!t]\centering
\includegraphics[width=\textwidth]{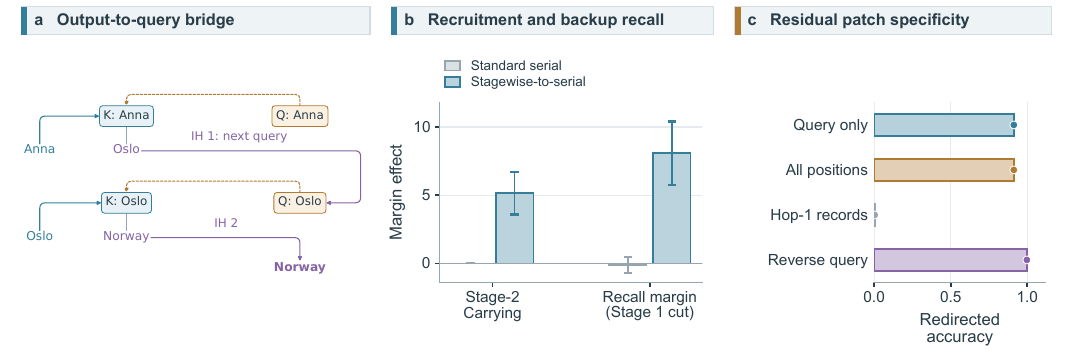}
\caption{\textbf{Stage availability and learned sequential use require separate tests.} a: The first retrieved value supplies the next query. b: The recruitment experiment measures later-stage Carrying and recall after the first stage is cut (means with sample SD). c: In the residual-stream two-hop experiment, a swapped donor redirects the endpoint when patched at the query or over the full sequence, but not at first-hop record positions. The latter are reported aggregate accuracies. Appendix Table~\ref{tab:recruit}A--D provides the complete controls.}
\label{fig:twoih}\label{fig:v32-circuit4}
\end{figure*}

The residual-stream experiment identifies a learned intermediate interface (Fig.~\ref{fig:twoih}; Appendix Table~\ref{tab:querycontrols}B,C). After a first-hop relation swap, patching only the first receiver's query-position residual redirects the endpoint prediction, whereas record-position patches do not. Receiver ablations separate the first stage's contribution to both lookups from the second stage's endpoint-specific role.

A separate positive-control suite explicitly reconstructs the second query from the first stage's predicted token distribution (Appendix Table~\ref{tab:querycontrols}A,D). In that circuit, competing-intermediate substitutions redirect the answer, and second-source access increases query-restoration gain (Appendix Table~\ref{tab:querycontrols}A,D). This control tests the dependency implemented by a designed query interface; Table~\ref{tab:twohop} evaluates learning that dependency in the residual stream.

\begin{table}[!ht]
\centering\small
\setlength{\tabcolsep}{3pt}
\caption{\textbf{Training routes in the residual-stream suite separate memorization and $\psi$-shortcuts from sequential composition.} Five-seed mean $\pm$ SD. Hop-1 uses full-vocabulary accuracy; unseen two-hop entries report candidate-pair accuracy with full-vocabulary accuracy in parentheses. Swap $\Delta$ measures the margin change after replacing the query intermediate.}\label{tab:twohop}\label{tab:twohop-full}
\fitwidth{\begin{tabular}{@{}lrrrr@{}}\toprule
Configuration & Hop-1 & 2-hop (table) & 2-hop (unseen) & Swap $\Delta$\\\midrule
1. $\text{IH}\times 1$ + 2-hop (fixed table) & $.966\pm.023$ & $1.000\pm.000$ & $.492\pm.034$ ($.000$) & $0.00\pm0.00$\\
2. $\text{IH}\times 1$ + stable $\psi$ (memory) & $.000\pm.000$ & $.980\pm.019$ & $.000$ ($\mathrm{R/ctlB}$)$^*$ & $0.00\pm0.00$\\
3. $\text{IH}\times 1$ + refresh ($\psi$-filter) & $.000\pm.000$ & $.653\pm.019$ & $.380$ ($\mathrm{ctlA}$)$^*$ & $0.00\pm0.00$\\
4. $\text{IH}\times 2$ (no 2-hop supervision) & $.916\pm.176$ & $.513\pm.105$ & $.488\pm.040$ ($.000$) & $0.05\pm0.28$\\
5. 2-hop from scratch (no IH) & $.000\pm.000$ & $1.000\pm.000$ & $.508\pm.055$ ($.508$) & $0.00\pm0.00$\\
6. Parallel readout ($\text{IH}_1(q)+\text{IH}_2(q)$) & $1.000\pm.000$ & --- & $.508\pm.044$ ($.045$) & $0.00\pm0.00$\\
\textbf{7. $\text{IH}\times 2$ + 2-hop (composed)} & $\mathbf{1.000\pm.000}$ & $\mathbf{1.000\pm.000}$ & $\mathbf{.980\pm.011}$ ($\mathbf{.977}$) & $\mathbf{19.15\pm2.47}$\\\bottomrule
\end{tabular}}
\par\smallskip\parbox{\linewidth}{\footnotesize $^*$In the value-consistency control grid ($\mathrm{Familiar}/\mathrm{Rebind}/\mathrm{ctlA}/\mathrm{ctlB}$), stable-$\psi$ training reaches $.980/.987$ on $\mathrm{Familiar}/\mathrm{ctlA}$ without reading record~2 (\texttt{Block R2}: $.980$) but drops to $.000$ on $\mathrm{Rebind}/\mathrm{ctlB}$; partial-refresh training hits $1.000$ on $\mathrm{Rebind}$ without reading record~1 (\texttt{Block R1}: $1.000$) by selecting the $\psi$-violating candidate, collapsing to $.380$ on the discriminator cell $\mathrm{ctlA}$.}
\end{table}

\subsection{Depth, Recurrence, and Intermediate Tokens}
\label{sec:statebench-icl}
Sequential retrieval can be executed across layers, recurrent passes, or generated intermediate tokens~\citep{dehghani2019universal,saunshi2025looped,wei2022cot}. We compare these routes on controlled state-tracking tasks inspired by StateBench's hierarchy~\citep{petty2026statebench} and on compositional in-context tasks. Training uses shorter assignment sequences; evaluation also tests longer sequences (Table~\ref{tab:statebench-icl-unification}). Appendix~\ref{app:semantic-statebench-protocols} specifies the task constructions and compute configurations.

On the tested state-tracking constructions, a single stage handles immutable initial lookup and latest-overwrite lookup, but performs poorly on group-accumulator and reference-assignment composition. Adding a second pass, a latent token, or a second stage with endpoint-only supervision leaves the harder tasks unresolved. Aligning the intermediate output with the next query supports strong generalization across all three execution routes. Longer assignment sequences test length transfer at the task's prescribed dependency structure. Across routes, intermediate-interface training is the shared experimental factor.

The compositional in-context tasks define custom arithmetic rules, category taxonomies, codebooks, and finite-state transitions within each prompt. A second stage with intermediate alignment succeeds across these task families, while single-stage and endpoint-only models remain close to the candidate-pair baseline (Table~\ref{tab:icl-composition}). Models without demonstrations additionally fail full-vocabulary selection. Together with the query-bridge interventions, these results connect composition to the learned use of an intermediate retrieval, rather than merely to the presence of extra layers.

\begin{table*}[!t]\centering\small\setlength{\tabcolsep}{3.5pt}
\caption{\textbf{Unifying single-pass depth ($\text{IH}\times 2$), weight-shared recurrence ($\text{IH}\times 1 + \text{Loop}$), and intermediate-token generation ($\text{IH}\times 1 + \text{CoT}$) across StateBench-inspired state tracking and compositional in-context learning (3 seeds: 42, 43, 44).} \textbf{A}: Mean accuracy across training lengths ($N{=}12,24$ assignments) and longer OOD sequences ($N{=}40,64$; up to $2.7\times$ training length). \textbf{B}: Full-vocabulary accuracy (mean $\pm$ SD) and candidate-pair accuracy (in parentheses) across four compositional ICL rule families.}
\label{tab:statebench-icl-unification}\label{tab:icl-composition}\label{tab:statebench-full}\label{tab:icl-full}
\fitwidth{\begin{tabular}{@{}llrrrrrrrr@{}}\toprule
\multicolumn{10}{@{}l}{\textbf{A. State-tracking accuracy across assignment lengths ($N\in\{12,24\}$ in-distribution; $N\in\{40,64\}$ length transfer)}}\\
 & & \multicolumn{2}{c}{Tier 1: $\mathcal R$-trivial ($k{=}1$)} & \multicolumn{2}{c}{Tier 2: Aperiodic ($k{=}1$)} & \multicolumn{2}{c}{Tier 3: Soluble ($k{=}2$)} & \multicolumn{2}{c}{Tier 4: General finite ($k{=}2$)}\\
\cmidrule(lr){3-4}\cmidrule(lr){5-6}\cmidrule(lr){7-8}\cmidrule(l){9-10}
Execution route & Supervision \& architecture & $N{=}12 / 24$ & $N{=}40 / 64$ & $N{=}12 / 24$ & $N{=}40 / 64$ & $N{=}12 / 24$ & $N{=}40 / 64$ & $N{=}12 / 24$ & $N{=}40 / 64$\\\midrule
No local shift & 1. Full attention (NoPE, 4L) & $.362 / .286$ & $.185 / .096$ & $.193 / .188$ & $.146 / .138$ & $.130 / .133$ & $.156 / .125$ & $.195 / .208$ & $.203 / .177$\\
One stage ($k{=}1$) & 2. $\text{IH}\times 1$ Single-Stage (2L) & $\mathbf{1.000 / 1.000}$ & $\mathbf{1.000 / 1.000}$ & $\mathbf{1.000 / 1.000}$ & $\mathbf{1.000 / 1.000}$ & $.281 / .263$ & $.250 / .258$ & $.357 / .398$ & $.406 / .385$\\\midrule
Endpoint-only ($k{=}2$) & 3. $\text{IH}\times 1 + \text{2-Step Loop}$ (2L shared) & $1.000 / 1.000$ & $1.000 / 1.000$ & $1.000 / 1.000$ & $1.000 / 1.000$ & $.263 / .260$ & $.258 / .203$ & $.417 / .375$ & $.344 / .312$\\
Endpoint-only ($k{=}2$) & 4. $\text{IH}\times 1 + \text{1-Token}$ (2L latent token) & $1.000 / 1.000$ & $1.000 / 1.000$ & $1.000 / 1.000$ & $1.000 / 1.000$ & $.255 / .221$ & $.271 / .266$ & $.375 / .354$ & $.375 / .352$\\
Endpoint-only ($k{=}2$) & 5. $\text{IH}\times 2$ Two-Stage (4L) & $1.000 / 1.000$ & $1.000 / 1.000$ & $1.000 / 1.000$ & $1.000 / 1.000$ & $.255 / .263$ & $.279 / .271$ & $.508 / .438$ & $.500 / .443$\\\midrule
\textbf{Aligned ($\mathbf{k{=}2}$)} & \textbf{6.} $\mathbf{IH\times 1 + \text{2-Step Loop}}$ \textbf{(2L shared)} & $\mathbf{1.000 / 1.000}$ & $\mathbf{1.000 / 1.000}$ & $\mathbf{1.000 / 1.000}$ & $\mathbf{1.000 / 1.000}$ & $\mathbf{.984 / .982}$ & $\mathbf{.982 / .979}$ & $\mathbf{.992 / .987}$ & $\mathbf{.979 / .987}$\\
\textbf{Aligned ($\mathbf{k{=}2}$)} & \textbf{7.} $\mathbf{IH\times 1 + \text{1-Token CoT}}$ \textbf{(2L)} & $\mathbf{1.000 / 1.000}$ & $\mathbf{1.000 / 1.000}$ & $\mathbf{1.000 / 1.000}$ & $\mathbf{1.000 / 1.000}$ & $\mathbf{1.000 / 1.000}$ & $\mathbf{1.000 / 1.000}$ & $\mathbf{1.000 / 1.000}$ & $\mathbf{1.000 / 1.000}$\\
\textbf{Aligned ($\mathbf{k{=}2}$)} & \textbf{8.} $\mathbf{IH\times 2}$ \textbf{Two-Stage (4L)} & $\mathbf{1.000 / 1.000}$ & $\mathbf{1.000 / 1.000}$ & $\mathbf{1.000 / 1.000}$ & $\mathbf{1.000 / 1.000}$ & $\mathbf{.990 / .977}$ & $\mathbf{.990 / .974}$ & $\mathbf{1.000 / 1.000}$ & $\mathbf{1.000 / 1.000}$\\\bottomrule
\end{tabular}}
\par\smallskip
\fitwidth{\begin{tabular}{@{}llrrrr@{}}\toprule
\multicolumn{6}{@{}l}{\textbf{B. Compositional in-context learning across rule families (full-vocabulary accuracy $\pm$ SD, with candidate-pair accuracy in parentheses)}}\\
Arm & Training \& architecture configuration & Arithmetic & Dynamic taxonomy & Codebook cipher & State transition\\\midrule
1. No demonstrations & Parametric only (4L $\text{IH}\times 2$, no context rules) & $.129\pm .029\ (.507)$ & $.245\pm .036\ (.466)$ & $.089\pm .005\ (.496)$ & $.118\pm .008\ (.499)$\\
2. One stage ($k{=}1$) & 2L hybrid with in-context demonstrations & $.504\pm .036\ (.509)$ & $.499\pm .032\ (.499)$ & $.549\pm .019\ (.551)$ & $.495\pm .032\ (.496)$\\
3. Two stages, endpoint-only & 4L hybrid ($k{=}2$) without intermediate alignment & $.464\pm .018\ (.464)$ & $.478\pm .014\ (.478)$ & $.494\pm .036\ (.494)$ & $.500\pm .048\ (.500)$\\
\textbf{4. Two stages, aligned} & \textbf{4L hybrid ($\mathbf{k{=}2}$) with intermediate alignment} & $\mathbf{.993\pm .006\ (.993)}$ & $\mathbf{.990\pm .006\ (.990)}$ & $\mathbf{.990\pm .010\ (.990)}$ & $\mathbf{.988\pm .000\ (.988)}$\\\bottomrule
\end{tabular}}
\end{table*}

\subsection{Transfer Through Continued Relational Training}
We also continue training the larger pretrained recurrent hybrid on a mixture of language-modeling data and synthetic relational chains (Appendix~\ref{app:transfer}). A single relational template fits its familiar order but transfers poorly to reversed layouts. Varying symbols and layouts improves recombination within the covered systems, while unseen notations and free-form prose remain difficult (Appendix Fig.~\ref{fig:appendix-transfer}). The mixture also increases code perplexity relative to language-only continuation (Appendix Table~\ref{tab:relational-ppl}).

These results suggest combining refreshed relational supervision with varied linguistic presentations and explicit intermediate alignment. The next training-design question is whether such mixtures can improve sequential use while preserving general language and code modeling, rather than merely fitting a larger family of templates.

\section{Discussion and Related Work}
\label{sec:discussion}

\paragraph{One interface, three learning problems}
The source key $K_{B_1}$ connects predecessor Carrying, contextual disambiguation, and sequential use. Address radius $r$ describes the context assembled before a lookup; retrieval depth $k$ describes how lookups depend on one another. This distinction explains why richer local context can resolve competing sources without adding a global retrieval, and why recruiting a second induction stage does not itself teach the model to use the first result as a new query. Hybrid blocks expose these interfaces clearly, while the functional probes also apply to full-attention and released models.

\paragraph{A unified perspective on efficient mixers, sparse attention, looped recurrence, and Chain-of-Thought in the $(r, k)$ framework}
Organizing contextual computation around the source-address interface $K_{B_1}=A_r(X)_{B_1}$ and the output-to-query bridge $q_{i+1}=\mathcal B(y_i)$ clarifies the distinct roles of modern architectural and inference-time mechanisms:
\begin{enumerate}[leftmargin=*,nosep]
\item \emph{Sliding-window attention (SWA) vs.\ linear/gated recurrence (DeltaNet/GDN) along the address-radius axis ($r$)}: Stateless sliding-window layers act as clean local shift operators that form $K_{B_1}$ rapidly because they do not share parameters with a persistent intra-layer state $M_t$ competing for \texttt{lag-0} write/read capacity (Appendix~\ref{app:swa-placement} and Table~\ref{tab:swa-and-placement}). However, a pre-receiver stack of $L$ SWA layers with window $W$ is strictly bounded by the local horizon $R_{\mathrm{local}}\le L(W{-}1)$ (Table~\ref{tab:swa} and Proposition~\ref{prop:locality}). Linear and gated delta-rule recurrences~\citep{schlag2021linear,yang2024gla,yang2025gdn} play a dual role: they propagate contextual cues beyond the fixed kernel horizon via persistent state $M_t$ and directly serve near-range or latest-association queries, but sharing short-convolution taps between intra-layer state processing (\texttt{lag-0}) and cross-layer Carrying (\texttt{lag-1}) creates route competition during formation unless decoupled by Q/V lag specialization, SoftPrior initialization, or tap isolation (Section~\ref{sec:formation}).
\item \emph{Sparse attention and conditional $n$-gram memory at the candidate-selection boundary}: Token-sparse attention mechanisms such as Native Sparse Attention (NSA~\citep{yuan2025nsa}) restrict global attention to a selected subset of historical positions. In our framework, sparse routing acts at the candidate-localization boundary ($P_{\mathrm{cand}}$) and must satisfy two requirements: first, block/token selection keys must already incorporate the local predecessor address $K_{B_1}$ so that the target source is included in the active candidate set; second, once multiple category-compatible candidates enter the sparse set, fine-grained source keys enriched with earlier context ($r\ge 2$) are required to raise conditional target selection $R_{\mathrm{in}}$ above the $50\%$ indistinguishability bound (Section~\ref{sec:context}). Conversely, conditional $n$-gram lookup modules such as Engram~\citep{cheng2026engram} can be understood as externalizing fixed-radius local composition ($S^r$) into an $O(1)$ table lookup: when routed into the historical value position before the receiver's key projection, an $n$-gram feature directly supplies a pre-composed multi-token address to $K_{B_1}$ without requiring $L$ sequential local relay layers.
\item \emph{Single-pass depth ($\text{IH}\times 2$), weight-shared Loops, and Chain-of-Thought (CoT) along the sequential-depth axis ($k$)}: Our sequential state-tracking comparison (Table~\ref{tab:statebench-icl-unification}) demonstrates that stacked two-stage hybrids ($\text{IH}\times 2$), weight-shared looped recurrence ($\text{IH}\times 1 + \text{Loop}$~\citep{dehghani2019universal,saunshi2025looped}), and autoregressive intermediate-token generation ($\text{IH}\times 1 + \text{CoT}$~\citep{wei2022cot}) are three physical realizations of the same sequential dependency $y_2=\mathcal R_2(\mathcal B(\mathcal R_1(q_1)))$. Specifically: (i) two-stage depth realizes $k{=}2$ \emph{spatially} across distinct layers within one forward pass via a hidden residual bridge ($\mathrm{L1}\to\mathrm{L3}$), trading parameter depth for single-token latency; (ii) looped recurrence realizes $k{=}2$ \emph{temporally} in latent space by reusing a single $\text{IH}\times 1$ stage across recurrent passes ($h_q^{(t)}\mapsto Q_q^{(t+1)}$), saving parameters while requiring subspace alignment between the stage's output and its own query projection; and (iii) CoT realizes $k{=}2$ \emph{autoregressively} in token space by projecting the intermediate retrieval $y_1$ through the language-model head into an explicit token and re-encoding it through the embedding and local-shift layers as the next query. Because CoT reuses the model's native token-input interface and next-token cross-entropy objective, it avoids hidden-subspace mismatch and achieves $1.000$ accuracy across all four state-tracking tiers once intermediate steps are supervised.
\end{enumerate}

\paragraph{Implications for training and experimental scope}
Our experiments separate address formation, stage recruitment, and bridge alignment:
\begin{enumerate}[leftmargin=*,nosep]
\item \emph{Guide address formation and stage placement, then train sequential use}: Shared-tap gradient interventions and norm-controlled initialization experiments identify ways to promote predecessor Carrying in controlled models (Table~\ref{tab:state}B,D). In three-seed paired pretraining comparisons at 79M and 318M (Appendix Table~\ref{tab:pretraining-full}), SoftPrior does not produce a statistically resolvable net gain in aggregate perplexity ($+1.65$ at 79M, $t{=}1.85$; $-1.95$ at 318M, $t{=}1.01$; both below the three-seed resolution thresholds of $3.85$ and $8.34$, where resolving a $1\text{--}2$ PPL net effect requires $8\text{--}10$ seeds per scale). Instead, SoftPrior consistently raises earlier-receiver Matching in $6/6$ seeds across both scales and reshapes the distance/domain profile at 79M (worsening adjacent/near recall in $3/3$ seeds while improving distant recall, distant disambiguation, and all three code-identifier reuse buckets in $3/3$ seeds; at 318M, the $3/3$ near-range cost and $3/3$ earlier-stage Carry/Match gains replicate, while distant/code bucket differences are mixed across seeds). Early predecessor-tap attenuation relocates the primary circuit to the later stage (Appendix Table~\ref{tab:pretraining-full}C), while stagewise-to-serial training preserves both stages simultaneously (Appendix Table~\ref{tab:recruit}A); intermediate-aligned supervision then trains the output-to-query dependency needed for sequential retrieval.
\item \emph{Natural-corpus shortcut dominance and sequential alignment}: The OpenWebText \texttt{L0..L6} ladder audit (\texttt{L6=0}, with all $6/6$ AND-gate survivors being verbatim repeated spans; Appendix Table~\ref{tab:corpus-ladder}E) and Python AST audit ($0/42$ conflict-free two-hop instances) explain why standard pretraining favors $\text{IH}\times 1$ plus parametric-memory or copy shortcuts. Achieving reusable $k{=}2$ generalization requires both $k\ge 2$ effective stages and intermediate-aligned supervision across stacked stages, loops, or CoT tokens (Tables~\ref{tab:twohop}--\ref{tab:statebench-icl-unification}), while continued training at 318M shows that synthetic relational gains must be balanced against unseen-notation transfer and code perplexity (Appendix Fig.~\ref{fig:appendix-transfer} and Table~\ref{tab:relational-ppl}F).
\item \emph{Scope and remaining boundaries}: Source-key restoration measures a functional receiver interface; pretrained source-window lesions establish checkpoint necessity, whereas the controlled branch experiments isolate developmental gradient interference. The single-read visibility theorem (Proposition~\ref{prop:indistinguishability}) applies under its specified access restrictions, and the corpus audits characterize the sampled candidate constructions.
\end{enumerate}

\paragraph{Relation to circuit and architecture studies}
Classical circuit analyses established predecessor-to-key composition and induction-based copying~\citep{elhage2021framework,olsson2022induction,singh2024induction}, while causal patching and emergence studies track these computations in uniform Transformers~\citep{geiger2023causal,zhang2024patching,chen2024unveiling,musat2025emergence,aoyama2026predicting}. Our block-update probes and $K_{B_1}$ restorations connect the same operations across heterogeneous local and global layers and isolate shared-tap competition with recurrent state readout. Whereas prior hybrid analyses focus on single-hop induction priors or expressivity~\citep{qiao2026rethinking,wang2025hybrid,cooper2026expressivity,merrill2026olmohybrid}, we extend the source-key interface to representation-driven category matching, distractor-key competition, and contextual disambiguation within $R_{\mathrm{local}}\le L(W{-}1)$. Finally, complementing recall benchmarks~\citep{arora2024zoology,jelassi2024copying,park2024mambaicl,arora2025mechanistic}, state-tracking hierarchies~\citep{petty2026statebench}, graph depth bounds~\citep{xu2019powerful,sanford2024understanding}, and looped/CoT reasoning~\citep{dehghani2019universal,saunshi2025looped,wei2022cot}, we isolate the intermediate output-to-query bridge that unifies spatial depth, recurrence, and CoT while separating sequential composition from memorization and single-hop shortcuts.

\section{Conclusion}
\label{sec:related-conclusion}
The token before the value supplies an address that training must form, enrich, and learn to use. Source-key interventions connect these processes across hybrid settings. Earlier context resolves competition within one lookup, while sequential retrieval requires an aligned output-to-query bridge across stacked stages, looped recurrence, or generated intermediate tokens. Address radius and retrieval depth therefore provide a common language for diagnosing circuit formation and designing training that supports their composition.

\paragraph{Code and disclosure}
Experiment code and raw evaluation logs are available at \url{https://github.com/ckpassenger/bind-match-copy}. Generative AI assisted manuscript revision, methodological feedback, and experimental-code generation; the authors take full responsibility for the text, code, results, and claims.

\FloatBarrier
\bibliographystyle{IEEEtranN}
\bibliography{references}
\FloatBarrier
\appendices
\section{Experimental Protocols}
\label{app:protocols}

\subsection{Shared Tasks and Reporting}
Controlled training cycles equally through long-range associative recall, short-range associative recall, contextual disambiguation, and latest-association recall. The first two require retrieving an associated continuation. Disambiguation holds the immediate anchor fixed and distinguishes sources using earlier context. Latest-association recall asks for the latest continuation of a repeated key. The dedicated two-hop task supplies chains $a_i\to b_i\to c_i$ and queries only $a_i$, with its own training and evaluation.

Controlled summaries use five training seeds, 42--46, with means and sample standard deviations across seeds. Comparisons use the same generated training batches within a seed when specified by the experiment script. Controlled evaluations use 128 prompts per task and seed. Pretrained recurrent-hybrid localization at both scales retains the supplied seed-level measurements; Qwen localization uses 30 prompts per layer in its export. The Qwen source-key restoration test uses 200 paired prompts. Restoration tables identify their uncertainty summaries separately. Captions distinguish prompt replication from independent training runs.

Patch loss is clean minus patched margin. Restoration gain is restored minus corrupted margin. Source-tap intervention tables report intervened minus clean margin, so disruption is negative there. Associative-recall accuracy in the controlled formation experiments is the fraction with positive target-minus-distractor margin. The two-hop comparison reports 128-way full-vocabulary top-one accuracy on Hop-1 ($a_1\to b_1$, chance $.008$) and both candidate forced-choice accuracy ($c_1$ vs.\ $c_2$, chance $.500$) and 128-way full-vocabulary accuracy on the second hop. Each experiment specifies its task, budget, and replicate unit.

\subsection{Controlled Memories and Local Paths}
Controlled models use width 128, four heads, and FFN width 256, with AdamW, learning rate $1.5\times10^{-3}$, weight decay $.01$, and batch size 32. For one head, their local and recurrent outputs combine as
\begin{equation}
 o_t=\frac{V_t+g\sum_{j<t}\gamma^{t-j}\beta_j(Q_t^\top K_j)V_j}
 {\sqrt{1+g^2/4}}.
 \label{eq:controlledmemory}
\end{equation}
Here $Q,K$ are normalized, short-convolved features and $\beta_j$ is a sigmoid write gate. Output projection, residual addition, and an FFN complete the layer. The state-strength sweep varies retention $\gamma$ and readout gain $g$ in one memory--attention pair for 1,400 steps. This controlled block is an additive gated memory; the pretrained DeltaNet blocks use a delta-rule state update.

The formation experiment uses three controlled memory layers followed by attention, with $\gamma=.90$, $g=1.5$, and 3,600 training steps. It changes the local-path configuration: default four taps, two taps, or hard Q/V masks. Two taps retain offsets zero and one. The hard split keeps only lag zero in Q/K/V of head 0; heads 1--3 retain only lag one in Q/V and lags zero/one in K. Masks apply during the forward pass and after optimization, with masked gradients zeroed. The figure shows all five seeds at steps 600, 1,000, 1,400, 2,000, 2,800, and 3,600.

To decompose shared-tap interference, the diagnostic evaluates each of the four tasks at every sampled checkpoint rather than inheriting the training cycle's current task. It separately backpropagates through the local and recurrent branches, records each tap's gradient inner product, and averages the checkpoint measurements within each seed. Table~\ref{tab:gradients} reports the resulting task-specific measurements.

\subsection{Pretrained and Released Models}\label{app:settings}
The two pretraining scales have 79M and 318M parameters. The smaller setting has eight layers, width 512, and eight heads; the larger has 16 layers and width 1,024. DeltaNet and SWA hybrids arrange efficient layers in groups of three before global attention. The SWA window comparison changes only $W=2,4,8,16$ at the smaller pretraining scale. The baseline sliding-window configuration uses window four.

Cross-scale initialization comparisons use three independent training seeds (42, 43, and 44) per condition at both 79M and 318M; baseline localization maps retain three training seeds, and the 79M early-stage attenuation run is evaluated at seed 42.

These checkpoints use GPT-2 tokenization, packed 1,024-token OpenWebText sequences, and AdamW with $(\beta_1,\beta_2)=(.9,.95)$ for 30K optimizer steps. The 79M recipe uses micro-batch 16 with gradient accumulation 4 (effective batch 64, $\approx 1.97\mathrm{B}$ tokens) and a 1K-step warmup; the 318M recipe uses micro-batch 32 with gradient accumulation 4 (effective batch 128, $\approx 3.93\mathrm{B}$ tokens) and a 2K-step warmup. Learning rates peak at $3\times10^{-4}$ and decay to $3\times10^{-5}$ by 30K steps; gradient clipping is 1.0. Comparisons are matched within each scale. Learned absolute position embeddings are present; SWA also applies RoPE to Q/K. The recurrent pretraining exports identify their efficient layers as DeltaNet. The released Qwen3-4B and Qwen3.5-4B checkpoints have 36 and 32 layers, respectively. All model families share the functional probe definitions.

\subsection{SoftPrior Initialization}
\label{app:softprior}
SoftPrior retains all four convolution taps as trainable parameters. Initial longer-offset taps are multiplied by $0.05$. One quarter of heads are biased toward current-token state processing by scaling the predecessor tap by $0.10$. In the remaining heads, current/predecessor scales are $0.10/1.20$ on the query path and $0.25/1.15$ on the value path. These are initialization coefficients, not forward masks. Hard-mask comparisons permanently constrain the selected taps, as specified above. Norm-matched isotropic controls match the initial Q/V filter norms without a directional lag bias. Swapped-lag controls move the predecessor boost to lag two. Norm-preserved SoftPrior rescales each head's four-tap filter back to its default norm after applying the directional bias; Table~\ref{tab:initialization-acquisition} reports the resulting norms.

\subsection{Stagewise-to-Serial Training Schedule}
\label{app:stageprotocol}
The recruitment comparison uses $[\mathrm{Memory},\mathrm{Attention}]^2$, batch 32, and 1,800 steps. Both arms share the same \textsc{SoftPrior} tap initialization and the same four-task sequence. The initialization biases local lag-one features while keeping all four taps trainable; it is held fixed across this comparison.

Let $x_0$ be the token embedding stream and $\Delta_1=h_1-x_0$ the first stage's update. For steps 1--600, each stage receives $x_0$ and is supervised on the full batch through the \emph{shared} output head, with loss $\frac12\mathcal L(h_1)+\frac12\mathcal L(h_2(x_0))$. For steps 601--1,100, define $\alpha=(t-600)/500$ and feed $x_0+\alpha\Delta_1$ to Stage 2. If $\Delta_2$ is its resulting update, the transition loss is
\begin{align}
 \mathcal L_t={}&(1-w)\mathcal L(x_0+\Delta_1+\Delta_2)\nonumber\\
 &+w\mathcal L(x_0+0.15\,\operatorname{sg}(\Delta_1)+\Delta_2),\label{eq:stageaux}\\
 w={}&0.5(1-\alpha),\nonumber
\end{align}
where $\operatorname{sg}$ stops gradients. The auxiliary term emphasizes Stage 2 while the serial connection is introduced. Steps 1,101--1,800 and all evaluations use the ordinary serial model and its standard loss. A controlled ablation compares a shared output head and gradual transition with independently initialized heads and an abrupt switch. Independent heads create competing demands on the shared embedding and leave later stages poorly aligned with the surviving readout. The shared-head transition preserves later-stage Carrying. The curriculum therefore coordinates the readout interface as well as recruiting each stage.

\subsection{Two-Hop Training and Query Interventions}
The residual-stream suite (Appendix Table~\ref{tab:twohop-full}) compares seven configurations across five seeds (42--46, batch size 32): fixed-table memorization, stable parametric $\psi$, partial-refresh filtering, two stages trained only on one-hop tasks, endpoint training from scratch, parallel readout, and two stages trained with refreshed intermediate and endpoint supervision. The value-consistency grid and record ablations distinguish contextual composition from parametric lookup and inconsistency filtering. Table~\ref{tab:querycontrols}B,C reports receiver ablations and L1$\to$L3 query-residual relay patches for this suite.

The explicit query-reconstruction positive control is a separate five-seed suite (Table~\ref{tab:querycontrols}A,D). Its second stage receives $x_0$ at prefix positions and $\operatorname{softmax}(z_1(q)/T)E+p_q$ at the query, with $T=.25$, token-embedding matrix $E$, and positional embedding $p_q$. The intermediate is predicted by Stage~1; the interface explicitly re-embeds that prediction. Training uses 900 steps with a weighted sum of hop-1 and hop-2 losses (.4/.6). The unbridged control instead passes $h_1$ directly to Stage~2 with endpoint supervision. Query swaps hold prefix records fixed. The restoration factorial crosses the supplied intermediate with second-source blocking inside attention; all four cells share the corrupted background, and blocking gives zero attention to that source.

\subsection{Semantic Retrieval and Sequential-Task Protocols}
\label{app:semantic-statebench-protocols}
The semantic extension and interference experiments (Section~\ref{sec:semantic-competition-disambig} and Table~\ref{tab:semantic-competition-disambig}) use a 4-layer $[\mathrm{SWA2},\mathrm{Full},\mathrm{SWA2},\mathrm{Full}]$ hybrid ($d{=}64$, 4 heads, $150{,}592$ parameters) across three independent seeds (42, 43, 44). The disambiguation experiment instead uses the single-read configuration specified below, with the same width and parameter count. All three experiments use full-vocabulary argmax accuracy:
\begin{itemize}[leftmargin=*,nosep]
\item \emph{Representation-only category extension}: Starting from literal-IH checkpoints, all attention, MLP, positional, output-head, and non-cue token weights are frozen; only the 16 adjective embedding rows ($16\times 64=1{,}024$ scalars) are trained for 3,000 steps (batch size 64, AdamW, $\text{lr}=2\times 10^{-3}$) on prompts where an unseen-in-prompt query adjective retrieves a randomly assigned label bound to same-category adjectives ($512$ test prompts per seed). Q/K tests patch projected activations rather than weights. The predecessor-edge intervention replaces the value content on that edge while retaining clean attention weights.
\item \emph{Category competition across context lengths}: Models are trained for 2,000 steps on exact adjective-to-color recall across 2, 4, 8, and 16 records (17, 29, 53, and 101 word-level tokens including BOS) and then adapted for 1,500 steps on category retrieval by updating only the 16 adjective embeddings. Evaluation ($256$ prompts per length and condition) compares cross-category distractors, $M_{\mathrm{conf}}\in\{1,3,7,15\}$ same-category conflicting distractors (both shared wrong code and up to 7 distinct wrong codes), and same-category same-answer distractors, alongside bidirectional $\mathrm{L1}$ distractor $K/V$ replacements.
\item \emph{Lag-$2/3/4$ identity disambiguation}: Inheriting the adapted embeddings in a 1-writer, 1-global-reader, 2-positionwise configuration where the final query reads only the four historical code positions ($29$ tokens, 4 records covering 2 person identities $\times$ 2 categories), models are warmed up for 600 steps and trained for 3,000 steps on mixed unique-candidate ($25\%$) and identity-disambiguated ($75\%$, half identical-word and half category-related unseen-in-prompt query word) prompts across identity lags $d\in\{2,3,4\}$. Evaluation uses $256$ prompts ($128$ identity-swapped twin pairs) per condition, testing source writer visibility windows $r\in\{1,2,3,4\}$, causal edge cuts (identity edge $t_i-d\to t_i$, \texttt{lag-1} word edge $t_i-1\to t_i$, and filler edges), and $\mathrm{L1}$ $K_{B_1}/V_{B_1}$ restorations.
\end{itemize}

The StateBench-inspired controlled state-tracking and compositional ICL suite (Section~\ref{sec:statebench-icl} and Table~\ref{tab:statebench-icl-unification}) trains and evaluates across seeds 42, 43, and 44 ($d{=}128$, 4 heads, batch size 32, AdamW $\text{lr}=1.5\times 10^{-3}$). For state tracking, models are trained on $N\in\{12,24\}$ state/entity assignments and evaluated on 128 prompts per tier at $N\in\{12,24\}$ (in-distribution) and $N\in\{40,64\}$ ($1.7\times$ and $2.7\times$ OOD length) across \texttt{r\_trivial} (first-assignment immutable lookup), \texttt{aperiodic} (latest-overwrite lookup), \texttt{soluble} ($\mathbb Z_2$/$\mathbb Z_3$ group-accumulator composition), and \texttt{general\_finite} (reference-assignment composition, $x\gets\mathrm{val}(y)$). The tier names organize these controlled constructions; sequence length is the number of assignments. The eight arms compare a 4-layer NoPE Transformer, a 2-layer single-stage $\text{IH}\times 1$ hybrid, and three $k{=}2$ paradigms---a 2-layer weight-shared hybrid looped for 2 recurrent steps ($\text{IH}\times 1 + \text{2-Step Loop}$), a 2-layer hybrid autoregressively generating 1 intermediate token before the answer ($\text{IH}\times 1 + \text{1-Token CoT}$), and a 4-layer 2-stage hybrid ($\text{IH}\times 2$)---each trained either with endpoint-only supervision (\texttt{Unaligned}) or with intermediate state/token alignment (\texttt{Composed}). The four compositional ICL tasks (\texttt{ICL1\_Custom\_Arithmetic}, \texttt{ICL2\_Dynamic\_Taxonomy}, \texttt{ICL3\_Codebook\_Cipher}, \texttt{ICL4\_FSM\_Transition}) supply per-prompt randomly generated two-stage rule demonstrations and report both 128-way full-vocabulary accuracy and two-candidate pair accuracy across 128 test prompts per seed.

\subsection{Task Names and Reproduction Keys}
\label{app:names}
The manuscript names tasks by the relation being tested. The source package provides the mapping to unchanged evaluation keys in its reproduction notes. A distance qualifier describes retrieval separation. Code identifier buckets use decoded-token heuristics and their original precedence rules; ``long-range'' describes the distance to a reused identifier, not the identifier's lexical length.

For natural-text evaluation, the distant recall bucket spans separations of 101--512 tokens; the distant-identifier rule uses separation greater than 100 tokens. The near recall bucket spans 1--20 tokens. ``Disambiguation'' denotes the shared-anchor task family; task accuracy and natural-text bucket perplexity remain separate metrics.

\section{Controls and Extended Readouts}
\label{app:controls}

\subsection{Head Localization and Probe Definitions}
\label{app:probe-definitions}\label{app:sourcekey-controls}
\begin{figure*}[!t]\centering
\includegraphics[width=174mm]{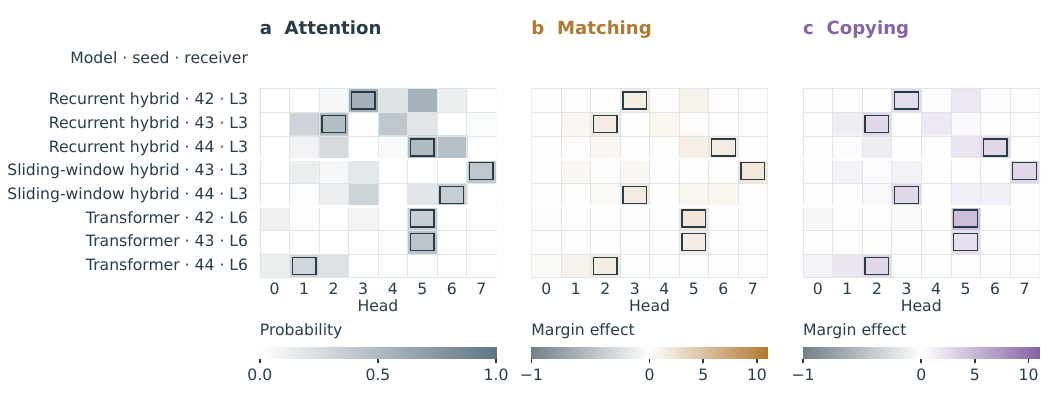}
\caption{\textbf{\matching\ and \copying\ identify the same dominant heads.} Rows align the same baseline model, seed, and receiver across attention probability, \matching, and \copying. The latter two use a shared signed margin scale. SWA includes seeds 43/44; Recurrent hybrids and Transformers include 42/43/44. Outlines mark row maxima.}
\label{fig:headcontrol}
\end{figure*}

We compare source attention with per-head \matching\ and \copying\ interventions across all eight baseline model--seed--receiver combinations in Fig.~\ref{fig:headcontrol}: the maximum \matching\ and \copying\ effects occur in the exact same attention head in every row, whereas raw attention weight maxima need not coincide. Carrying interventions change a predecessor ($\tau{=}1$) or an earlier cue ($\tau\ge 2$) and patch the historical value position $s{=}B_1$; Matching swaps historical associations and patches the receiver query $q$; Copying replaces the value-readout contribution while holding keys and attention weights fixed. Controlled stage-level Carrying is measured at the local layer immediately preceding each receiver, whereas pretraining summaries use block Carrying $C_{\mathcal B}=\sum_{\ell\in\mathcal B}\max(C_\ell,0)$ over the local layers preceding L3/L7 at 79M ($\mathrm{L0\text{--}L2}/\mathrm{L4\text{--}L6}$) and L7/L11 at 318M ($\mathrm{L4\text{--}L6}/\mathrm{L8\text{--}L10}$).

Source-key specificity controls are reported separately in Table~\ref{tab:sourcekey-specificity}, preserving a common model/interface order with the main restoration comparison.

\begin{table*}[!t]\centering\small\setlength{\tabcolsep}{6pt}
\caption{\textbf{Specificity controls for source-key restoration.} These interventions use the same corrupted backgrounds as Table~\ref{tab:sourcekey}. Columns report restoration through the source value, an off-site key, or the query. Controlled and 79M means and sample SD are retained as supplied; Qwen entries are reported point estimates. A dash denotes an unreported intervention.}
\label{tab:sourcekey-specificity}\label{tab:sourcekey-controls}
\begin{tabularx}{\textwidth}{@{}Xlrrr@{}}\toprule
Model & Interface & Source value $V_{B_1}$ & Off-site key $K_p$ & Query $Q_q$\\\midrule
\multicolumn{5}{@{}l}{\textit{Margin change}}\\
Controlled pair & L0$\to$L1 & $+0.12\pm0.05$ & $-1.10\pm0.42$ & $0.00\pm0.00$\\
Controlled $3{:}1$ block & L2$\to$L3 & $+0.17\pm0.06$ & $-1.25\pm0.49$ & $0.00\pm0.00$\\
Recurrent hybrid (79M) & L2$\to$L3 & $\le +0.06$ & $-1.34\pm0.74$ & $0.00\pm0.00$\\
Recurrent hybrid (318M) & L7, L11$^\dagger$ & $0.00$ & $+0.01$ & $0.00$\\
Qwen3.5-4B & L18$\to$L19 & $+0.03$ & --- & $+0.01$\\
\midrule
\multicolumn{5}{@{}l}{\textit{NLL change ($\Delta\mathrm{NLL}$)}}\\
SWA hybrid (79M, $W{=}2$) & L2$\to$L3 & $0.00\pm0.00$ & $-0.03\pm0.06$ & $0.00\pm0.00$\\
SWA hybrid (79M, $W{=}8$) & L2$\to$L3 & $0.00\pm0.00$ & $-0.01\pm0.04$ & $0.00\pm0.00$\\
\bottomrule
\end{tabularx}
\par\smallskip\parbox{\textwidth}{\footnotesize Controlled models use five training seeds and the 79M pretrained models use three. The Qwen3.5-4B key-restoration result uses 200 paired prompts with 95\% bootstrap intervals. $^\dagger$After L6 disruption, source-key gains at L7 and L11 are $1.33$ and $0.51$. A separate 318M single-receiver test at L2$\to$L3 gives source-key $\Delta\mathrm{NLL}=0.43$ against disruption $0.53$ (80.5\% recovery), with source-value recovery $-0.002$ and off-site-key recovery $+0.012$.}
\end{table*}

\subsection{Formation Timing, Shared-Tap Gradients, and Local Structure}
\label{app:formation-timing}\label{app:grad}\label{app:swa-placement}
\begin{figure*}[!t]\centering
\includegraphics[width=\textwidth]{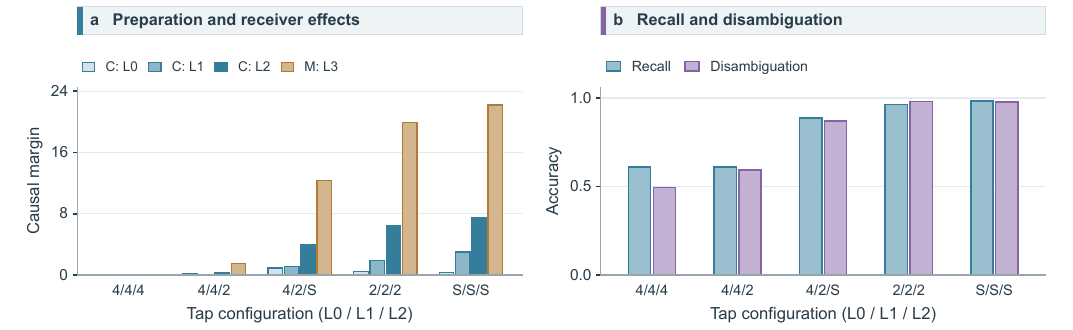}
\caption{\textbf{Local tap constraints align preparation, matching, and retrieval.} The two panels use the same configuration order. a: Carrying at local layers L0--L2 and Matching at receiver L3, shown as separate causal effects. b: Recall and disambiguation accuracy. Bars show the reported five-seed means at the controlled training endpoint. Numeric tap entries denote convolution width; S denotes the hard Q/V lag split. Acquisition counts and capped steps are retained in Table~\ref{tab:state}H.}
\label{fig:appendix-taps}
\end{figure*}

Table~\ref{tab:state} consolidates our eight formation and local-structure control suites:
\begin{enumerate}[leftmargin=*,nosep]
\item \emph{Trajectory onset and state-strength route shift (Table~\ref{tab:state}A)}: In pretraining trajectories (Fig.~\ref{fig:early-formation}), each run's final peak Carrying and Matching layers are traced backward through checkpoints. In the controlled memory--attention pair (Table~\ref{tab:state}A), increasing state retention and readout gain shifts recall from cross-layer Carrying ($7.07\to 0.00$) to intra-layer query state readout ($0.00\to 24.67$).
\item \emph{Filter-norm controls and shared-tap decoupling (Table~\ref{tab:state}B--D)}: Both standard SoftPrior and norm-preserved SoftPrior reach the recall threshold in $5/5$ seeds and outperform default, isotropic norm-matched, and swapped-lag controls (Panel~B). Task-specific branch-gradient decompositions confirm negative tap alignments and pre-formation \texttt{lag-0} pull from non-source state readouts (Panel~C), while gradient orthogonalization and tap/head isolation restore $5/5$-seed emergence and double Carrying magnitude in the strong-state regime (Panel~D).
\item \emph{Pretrained source-window lesions and local-mixer structure (Table~\ref{tab:state}E--H)}: Removing query-path history at the source window disrupts lag-2 disambiguation at both 79M and 318M whereas removing value-path history has minimal effect (Panel~E), and SoftPrior improves single-checkpoint latest-association and disambiguation margins (Panel~F). Under matched 1,500-step training, stateless SWA ($W{=}2$) reaches $1.000$ immutable and $.935$ overwrite accuracy with $+4.59\pm 1.34$ $\mathrm{write\to query}$ causal recovery because its shift path does not compete with a persistent state $M_t$ (Panel~G). Within a 3-layer pre-receiver block ($\mathrm{L0\text{--}L2}\to\mathrm{L3}$; Table~\ref{tab:state}H and Fig.~\ref{fig:appendix-taps}), restricting only $\mathrm{L2}$ fails ($1/5$), whereas progressive funneling ($\mathrm{4/2/split}$, $4/5$) or uniform short/split taps ($5/5$) concentrate Carrying monotonically toward $\mathrm{L2}$.
\end{enumerate}

\begin{table*}[!t]
\caption{\textbf{Formation, shared-tap interference, and local-structure controls.} Entries report multi-seed means $\pm$ sample SD in their respective controlled or pretraining settings. \textbf{A}: State retention and readout gain in a 2-layer memory--attention pair (1,400 steps, 5 seeds). \textbf{B}: Directional initialization and norm controls in 4-layer $3{:}1$ blocks (1,600 steps, 5 seeds; failures capped at budget). \textbf{C,D}: Shared-tap branch gradient alignment and direct decoupling in the strong-state regime ($\gamma{=}.95, g{=}2.0$, 5 seeds). \textbf{E,F}: Pretrained source-window history lesions (79M/318M) and single-checkpoint local-path comparisons (79M). \textbf{G,H}: Early 1,500-step retrieval formation across mixer families (3 seeds) and acquisition under layerwise tap constraints inside a $3{:}1$ local block (2,200 steps, 5 seeds). Fig.~\ref{fig:appendix-taps} shows the corresponding layerwise circuit and behavioral readouts.}
\label{tab:state}\label{tab:initialization-acquisition}\label{tab:gradients}\label{tab:sourcetaps}\label{tab:pretraining-local-controls}\label{tab:swa-and-placement}\label{tab:layerwise-taps}
\centering\small\setlength{\tabcolsep}{3pt}
\begin{minipage}[t]{.47\textwidth}\centering
\textbf{A. State retention and readout gain (5 seeds)}\par\smallskip
\fitwidth{\begin{tabular}{@{}lrrr@{}}\toprule
$(\gamma,g)$ & Recall acc. & Carrying & State readout\\\midrule
$(.50,.5)$ & $0.930\pm0.038$ & $7.07\pm1.94$ & $0.00\pm0.00$\\
$(.85,1.0)$ & $0.948\pm0.032$ & $8.24\pm1.40$ & $0.00\pm0.00$\\
$(.95,2.0)$ & $0.772\pm0.193$ & $3.80\pm2.51$ & $0.08\pm0.06$\\
$(.99,3.5)$ & $0.977\pm0.048$ & $0.00\pm0.01$ & $24.67\pm4.25$\\\bottomrule
\end{tabular}}
\par\medskip
\textbf{C. Shared-tap branch gradient alignment}\par\smallskip
\fitwidth{\begin{tabular}{@{}lrrr@{}}\toprule
Task & Neg.\ taps (/4) & Lag-1 cos. & Pre-form.\ pull ($\times 10^{-3}$)\\\midrule
Far recall & $1.68\pm0.60$ & $0.147\pm0.233$ & $-11.6\pm7.9$ (non-$B_1$)\\
Near recall & $2.08\pm0.56$ & $0.150\pm0.139$ & $-24.8\pm10.3$ (non-$B_1$)\\
Repeated anchors & $1.75\pm0.36$ & $0.033\pm0.054$ & $+23.6\pm11.5$ (at $B_1$)\\
State overwrite & $1.43\pm0.47$ & $0.243\pm0.182$ & $-26.1\pm11.2$ (non-$B_1$)\\\bottomrule
\end{tabular}}
\end{minipage}\hfill
\begin{minipage}[t]{.51\textwidth}\centering
\textbf{B. Directional initialization and filter-norm controls (5 seeds)}\par\smallskip
\fitwidth{\begin{tabular}{@{}lrrrr@{}}\toprule
Initialization (Q/V norm) & Reached & Capped step & Carrying & Comp.\ margin\\\midrule
Default ($6.46/6.57$) & 0/5 & $1600\pm0$ & $0.01\pm0.01$ & $1.91\pm1.06$\\
Isotropic norm ($3.77/3.75$) & 3/5 & $1460\pm219$ & $2.62\pm1.75$ & $4.64\pm2.51$\\
Swapped lag ($3.70/3.80$) & 1/5 & $1500\pm224$ & $1.28\pm1.50$ & $2.69\pm1.93$\\
ShortConv2 (hard mask) & 5/5 & $1240\pm152$ & $7.88\pm1.66$ & $7.03\pm0.98$\\
Q/V split (hard mask) & 5/5 & $940\pm195$ & $8.96\pm1.50$ & $8.02\pm1.24$\\
\textbf{SoftPrior (norm-pres.)} & \textbf{5/5} & $1060\pm230$ & $5.17\pm1.13$ & $7.46\pm1.65$\\
\textbf{SoftPrior (standard)} & \textbf{5/5} & $\mathbf{860\pm134}$ & $\mathbf{10.57\pm1.31}$ & $\mathbf{9.05\pm0.27}$\\\bottomrule
\end{tabular}}
\par\medskip
\textbf{D. Direct decoupling in Strong state ($\gamma{=}0.95, g{=}2.0$)}\par\smallskip
\fitwidth{\begin{tabular}{@{}lrrr@{}}\toprule
Intervention & Escaped & Carrying & Far accuracy\\\midrule
Shared-tap baseline & $3/5$ & $4.11\pm2.79$ & $.797\pm.204$\\
Gradient orthogonalization & $\mathbf{5/5}$ & $\mathbf{7.92\pm1.73}$ & $\mathbf{.949\pm.046}$\\
Tap/head isolation (norm-matched) & $\mathbf{5/5}$ & $\mathbf{8.90\pm1.37}$ & $\mathbf{.963\pm.046}$\\\bottomrule
\end{tabular}}
\end{minipage}
\par\medskip
\begin{minipage}[t]{.47\textwidth}\centering
\textbf{E. Pretrained lag-2 source-window lesions}\par\smallskip
\fitwidth{\begin{tabular}{@{}lrr@{}}\toprule
Source-window operation & 79M & 318M\\\midrule
Remove Q/K/V history & $-3.31$ & $-3.32$\\
Keep only lag 0--1 & $-.43$ & $+.04$\\
Remove V history & $+.11$ & $-.17$\\
Remove Q history & $-3.46$ & $-3.31$\\\bottomrule
\end{tabular}}
\end{minipage}\hfill
\begin{minipage}[t]{.51\textwidth}\centering
\textbf{F. Pretrained 79M local-path task margins (seed 42)}\par\smallskip
\fitwidth{\begin{tabular}{@{}lrr@{}}\toprule
Local path configuration & Latest association & Disambiguation\\\midrule
Default ($k{=}4$) & $0.94$ & $4.05$\\
Short convolution (\texttt{lag 0--1}) & $0.14$ & $4.36$\\
Q/V lag split & $0.44$ & $3.77$\\
SoftPrior initialization & $\mathbf{1.87}$ & $\mathbf{5.65}$\\\bottomrule
\end{tabular}}
\end{minipage}
\par\medskip

\begin{minipage}[t]{.48\textwidth}\centering
\textbf{G. Early retrieval across local mixers}\par\smallskip
\fitwidth{\begin{tabular}{@{}lrrrrr@{}}\toprule
Architecture & Immut. & Overwr. & Ref. & Write$\to$Q & Q$\to$src\\\midrule
SWA ($W{=}2$) & $\mathbf{1.000}$ & $\mathbf{.935}$ & $.659$ & $\mathbf{+4.59\pm1.34}$ & $-0.02$\\
Recurrent hybrid & $.997$ & $.883$ & $.674$ & $+3.58\pm1.10$ & $0.00$\\
Full attn (NoPE) & $.250$ & $.516$ & $.299$ & $0.00\pm0.00$ & $0.00$\\\bottomrule
\end{tabular}}
\end{minipage}\hfill
\begin{minipage}[t]{.50\textwidth}\centering
\textbf{H. Acquisition under layerwise tap constraints}\par\smallskip
\fitwidth{\begin{tabular}{@{}lrr@{}}\toprule
Taps ($\mathrm{L0/L1/L2}$) & Reached & Capped step\\\midrule
4 / 4 / 4 & $0/5$ & $2200$\\
4 / 4 / 2 & $1/5$ & $2180$\\
4 / 2 / split & $4/5$ & $1820$\\
2 / 2 / 2 & $5/5$ & $1240$\\
split / split / split & $5/5$ & $940$\\\bottomrule
\end{tabular}}
\end{minipage}
\end{table*}

\subsection{Three-Seed Pretraining Readouts and Stage Placement}
\label{app:pretraining-readouts}
Table~\ref{tab:pretraining-full} reports the three-seed paired comparison between Default and SoftPrior at 79M and 318M (seeds 42, 43, 44) alongside the stage-placement interventions:
\begin{enumerate}[leftmargin=*,nosep]
\item \emph{Seed-paired variance and the near-for-far trade-off (Table~\ref{tab:pretraining-full}A,B)}: Pairing by training seed reduces between-seed residual SD on overall perplexity by $\approx 30\%$ ($2.25\to 1.55$ at 79M and $4.80\to 3.36$ at 318M; an independent repeat of 79M Default at seed 42, $M_1$, gives $39.50$ vs.\ $39.54$ for $P_0$, confirming that seed variation dominates run noise). Because three seeds resolve minimum total-PPL differences of $3.85$ (79M) and $8.34$ (318M), neither scale exhibits a statistically significant net change in overall perplexity ($+1.65$ at 79M, $t{=}1.85$; $-1.95$ at 318M, $t{=}1.01$). Instead, two effects replicate across all $6/6$ seeds (both scales $\times$ 3 seeds): (i) earlier-receiver Matching rises in $6/6$ seeds ($+0.86$ mean at 79M L3, $+1.25$ at 318M L7), and (ii) near-range recall perplexity (\texttt{B3c\_near}, \texttt{B3c\_mid}) worsens in $6/6$ seeds. At 79M, this near-range cost pairs with a $3/3$-seed improvement across distant recall (\texttt{B3c\_far}), distant disambiguation (\texttt{B3d\_far}), and all three code-identifier reuse buckets (\texttt{code\_var\_reuse}, \texttt{code\_scope\_shadow}, and \texttt{code\_long\_id}, which drops by $-7.68, -14.37, -21.56$). At 318M, earlier-block Carrying also increases in $3/3$ seeds ($+0.83$), while distant/code bucket differences are mixed ($2/3$ seeds improved, reversed in seed 44). Later-block changes under SoftPrior are small and mixed across seeds ($2/3$ negative).
\item \emph{Steering stage placement via early \texttt{lag-1} attenuation (Table~\ref{tab:pretraining-full}C)}: Temporarily damping first-stage \texttt{lag-1} taps during the first 750 controlled steps or 3K pretraining steps prevents Stage~1 from absorbing the single-hop residual and relocates the primary circuit to Stage~2 (at 79M seed 42, Later Carrying jumps from $0.04\to 6.89$ and L7 Matching from $2.72\to 9.56$, lowering disambiguation PPL to $25.79$ and distant-ID PPL to $28.55$).
\end{enumerate}

\begin{table*}[!t]
\caption{\textbf{Three-seed pretraining readouts, seed-paired differences, and stage placement across parameter scales.} \textbf{A,B}: Default and SoftPrior use three matched seeds (42, 43, 44) at both 79M and 318M on the shared fixed evaluation set ($3{,}069{,}000$ tokens; overlapping bucket counts sum to $4.53\mathrm{M}$, so buckets are evaluated independently). Across $6/6$ seeds (both scales $\times$ 3 seeds), SoftPrior increases earlier-receiver Matching and worsens near-range recall (\texttt{Near recall}), while at 79M it improves \texttt{Far recall}, \texttt{Disambig.}, \texttt{Distant IDs}, and \texttt{Local IDs} in all $3/3$ seeds without a statistically resolvable net change in overall perplexity ($+1.65$ at 79M, $t{=}1.85$; $-1.95$ at 318M, $t{=}1.01$). \textbf{C,D}: Controlling first-stage \texttt{lag-1} taps during early training relocates the circuit to Stage~2 ($\mathrm{L6}\to\mathrm{L7}$ at 79M), whereas stagewise-to-serial training preserves both stages simultaneously.}
\label{tab:pretraining-full}\label{tab:softprior-cross-scale}\label{tab:v28-recipes-main}\label{tab:stage-placement}
\centering\small\setlength{\tabcolsep}{2.8pt}
\textbf{A. Language-modeling perplexity across buckets (3-seed mean $\pm$ SD and seed-paired differences $\mathrm{SP}-\mathrm{Def}$)}\par\smallskip
\fitwidth{\begin{tabular}{@{}llrrrrrrrr@{}}\toprule
Scale & Condition / Summary & Overall & New anchor & New cont. & Near recall & Far recall & Disambig. & Distant IDs & Local IDs\\\midrule
79M & Default ($P_0$, 3 seeds) & $41.68\pm2.25$ & $76.50\pm0.59$ & $372.00\pm8.80$ & $6.53\pm2.49$ & $2.60\pm0.04$ & $29.76\pm1.39$ & $49.93\pm3.04$ & $12.54\pm0.90$\\
 & SoftPrior (3 seeds) & $43.33\pm2.59$ & $75.75\pm0.36$ & $363.53\pm3.98$ & $9.58\pm3.13$ & $\mathbf{2.55\pm0.07}$ & $\mathbf{27.17\pm0.64}$ & $\mathbf{35.39\pm4.48}$ & $\mathbf{10.39\pm0.67}$\\
 & \emph{Paired $\Delta$ (s42 / s43 / s44)} & $+0.81/+3.44/+0.71$ & $-0.87/-1.64/+0.25$ & $-5.1/-21.5/+1.2$ & $+1.32/+6.19/+1.65$ & $\mathbf{-0.09/-0.03/-0.04}$ & $\mathbf{-0.92/-3.48/-3.36}$ & $\mathbf{-7.68/-14.4/-21.6}$ & $\mathbf{-1.21/-1.51/-3.73}$\\\midrule
318M & Default ($M_1$, 3 seeds) & $29.63\pm4.80$ & $49.51\pm12.91$ & $234.88\pm71.84$ & $5.99\pm0.88$ & $2.13\pm0.19$ & $17.21\pm3.73$ & $19.87\pm6.54$ & $7.22\pm2.31$\\
 & SoftPrior (3 seeds) & $27.68\pm1.56$ & $41.93\pm0.41$ & $193.80\pm4.96$ & $9.02\pm2.60$ & $2.03\pm0.05$ & $15.53\pm0.54$ & $17.36\pm1.41$ & $6.01\pm0.22$\\
 & \emph{Paired $\Delta$ (s42 / s43 / s44)} & $-5.75/-0.74/+0.63$ & $-22.1/-0.96/+0.28$ & $-117.9/-10.1/+4.7$ & $+6.96/+1.19/+0.93$ & $-0.32/-0.02/+0.02$ & $-5.08/-1.31/+1.35$ & $-7.42/-2.89/+2.80$ & $-3.31/-1.20/+0.87$\\\bottomrule
\end{tabular}}
\par\medskip
\begin{minipage}[t]{.48\textwidth}\centering
\textbf{B. Pretraining block Carrying and receiver Matching}\par\smallskip
\fitwidth{\begin{tabular}{@{}llrrrr@{}}\toprule
 & & \multicolumn{2}{c}{Earlier stage ($\mathrm{L3/L7}$)} & \multicolumn{2}{c}{Later stage ($\mathrm{L7/L11}$)}\\
\cmidrule(lr){3-4}\cmidrule(l){5-6}
Scale & Setting & Carry & Match & Carry & Match\\\midrule
79M & Default & $3.19\pm0.84$ & $5.39\pm1.50$ & $0.13\pm0.10$ & $2.93\pm0.27$\\
 & SoftPrior & $3.78\pm0.46$ & $\mathbf{6.25\pm0.74}$ & $0.07\pm0.05$ & $2.80\pm0.15$\\
 & \emph{Paired $\Delta$} & $+.97/-.10/+.90$ & $\mathbf{+1.7/+.29/+.57}$ & $-.20/-.03/+.04$ & $+.11/-.41/-.07$\\\midrule
318M & Default & $2.27\pm0.61$ & $0.79\pm0.37$ & $2.73\pm0.71$ & $5.44\pm1.58$\\
 & SoftPrior & $\mathbf{3.10\pm0.49}$ & $\mathbf{2.04\pm0.68}$ & $2.08\pm0.94$ & $4.80\pm1.31$\\
 & \emph{Paired $\Delta$} & $\mathbf{+1.6/+.26/+.66}$ & $\mathbf{+2.2/+1.3/+.27}$ & $-1.6/-1.5/+1.1$ & $-.66/-2.1/+.80$\\\bottomrule
\end{tabular}}
\end{minipage}\hfill
\begin{minipage}[t]{.50\textwidth}\centering
\textbf{C. Controlled \& pretrained stage placement (s42 matched)}\par\smallskip
\fitwidth{\begin{tabular}{@{}lrrr@{}}\toprule
Controlled 4L (5 seeds) & SoftPrior & Early atten. & Stagewise-serial\\\midrule
Stage-1 Carry / Match & $9.79\ /\ 20.12$ & $7.08\ /\ 4.32$ & $\mathbf{8.90\ /\ 18.45}$\\
Stage-2 Carry / Match & $0.00\ /\ 0.41$ & $\mathbf{0.54\ /\ 2.59}$ & $\mathbf{6.55\ /\ 14.12}$\\
Recall / Ambig.\ acc. & $.981\ /\ .967$ & $.908\ /\ .869$ & $\mathbf{.991\ /\ .984}$\\\midrule
Pretrained 79M (seed 42) & Default ($M_1$) & SoftPrior & + Early atten.\\\midrule
Earlier Carry / L3 Match & $2.31\ /\ 3.68$ & $\mathbf{3.27\ /\ 5.41}$ & $0.19\ /\ 0.02$\\
Later Carry / L7 Match & $0.25\ /\ 2.60$ & $0.04\ /\ 2.72$ & $\mathbf{6.89\ /\ 9.56}$\\
Far recall PPL & $2.59$ & $\mathbf{2.49}$ & $2.68$\\
Disambig.\ / Distant-ID PPL & $28.71\ /\ 47.98$ & $27.43\ /\ 38.74$ & $\mathbf{25.79\ /\ 28.55}$\\\bottomrule
\end{tabular}}
\end{minipage}
\end{table*}

\subsection{Semantic Competition, Visibility, and Local-Horizon Relays}
\label{app:semantic-details}\label{app:semantic-literal}
\begin{figure*}[!t]\centering
\includegraphics[width=152mm]{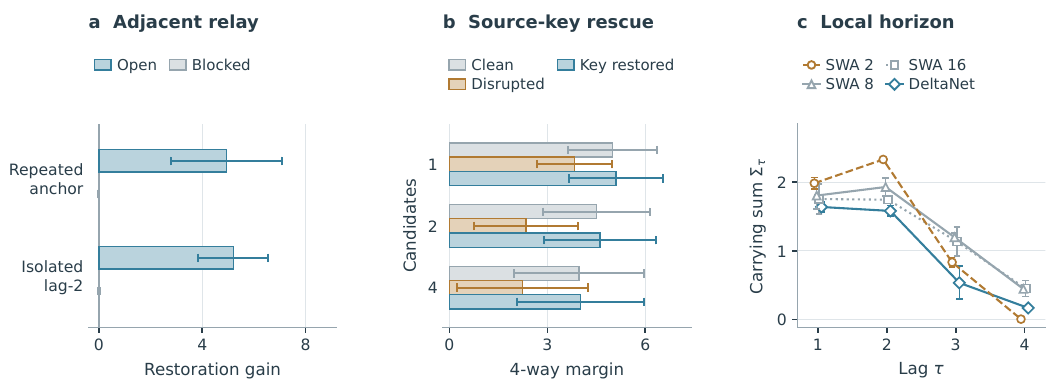}
\caption{\textbf{Extended relay and lag-spectrum controls.} a: Same-background anchor restoration crossed with open or blocked adjacent transfer. b: Context disruption and source-key restoration across candidate-set sizes ($K\in\{2,3,4\}$). c: Positive layer-effect sums compare lag profiles in 79M recurrent and sliding-window hybrids.}
\label{fig:relay-full}
\end{figure*}

Before and after cue-embedding category adaptation, the 4-layer checkpoints achieve literal/category accuracies of $100.0\pm0.0\%$/$36.1\pm5.0\%$ and $87.4\pm2.7\%$/$100.0\pm0.0\%$ (3 seeds), showing that broadening matching to categories slightly reduces exact-word selectivity on literal prompts. Table~\ref{tab:semantic-lengths} compiles the numerical readouts for semantic competition across lengths and conflict counts (Panel~A), bidirectional $\mathrm{L1}$ distractor $K/V$ exchanges (Panel~B), the single-read writer visibility matrix and full-window edge cuts/restorations (Panels~C,D), and multi-layer adjacent-relay restoration and SWA horizon cutoffs (Panels~E,F), complemented by the candidate-set size and cross-architecture lag spectrum profiles in Fig.~\ref{fig:relay-full}.

\begin{table*}[!t]
\caption{\textbf{Semantic competition, contextual disambiguation, and local-horizon relay controls.} \textbf{A,B}: Multi-layer category-adapted circuit (3 seeds); exact-recall accuracy (\%) across lengths and conflicting-record counts, plus bidirectional $\mathrm{L1}$ distractor $K/V$ exchanges at 101 tokens. \textbf{C,D}: Single-read writer (3 seeds); accuracy (\%) across source visibility windows $r\in\{1..4\}$ and full-window causal edge cuts/restorations. \textbf{E,F}: Multi-layer $2\times 2$ adjacent-relay restoration (5 seeds) and 79M SWA pre-receiver horizon profile (3 seeds).}
\label{tab:semantic-lengths}\label{tab:semantic-keys}\label{tab:context-access}\label{tab:context-cuts}\label{tab:relay}\label{tab:swa}
\centering\small\setlength{\tabcolsep}{3.2pt}
\begin{minipage}[t]{.51\textwidth}\centering
\textbf{A. Length vs.\ conflicting records (3 seeds, \%)}\par\smallskip
\fitwidth{\begin{tabular}{@{}rrrrrr@{}}\toprule
Rec. & Tok. & $M_{\mathrm{conf}}$ & Cross-cat. & Same-cat.\ conflict & Same ans.\\\midrule
2 & 17 & 1 & $100.0\pm0.0$ & $81.4\pm7.0$ & $100.0\pm0.0$\\
4 & 29 & 3 & $100.0\pm0.0$ & $55.7\pm10.7$ & $100.0\pm0.0$\\
8 & 53 & 7 & $100.0\pm0.0$ & $35.0\pm14.4$ & $100.0\pm0.0$\\
16 & 101 & 15 & $100.0\pm0.0$ & $20.8\pm14.4^\dagger$ & $100.0\pm0.0$\\
16 & 101 & 1 & $100.0\pm0.0$ & $79.8\pm8.4$ & $100.0\pm0.0$\\\bottomrule
\end{tabular}}
\end{minipage}\hfill
\begin{minipage}[t]{.47\textwidth}\centering
\textbf{B. Distractor $K/V$ exchange at 101 tokens (\%)}\par\smallskip
\fitwidth{\begin{tabular}{@{}llrr@{}}\toprule
Recipient input & L1 distractor patch & Accuracy & Shared err.\\\midrule
Conflicting records & None & $20.8\pm14.4$ & 78.8\\
Conflicting records & Cross-category $K$ & $\mathbf{100.0\pm0.0}$ & 0.0\\
Conflicting records & Cross-category $V$ & $21.0\pm13.9$ & 78.6\\
Cross-category records & Conflicting $K$ & $\mathbf{21.1\pm14.0}$ & 78.5\\
Cross-category records & Conflicting $V$ & $100.0\pm0.0$ & 0.0\\\bottomrule
\end{tabular}}
\end{minipage}
\par\medskip
\begin{minipage}[t]{.46\textwidth}\centering
\textbf{C. Identity visibility matrix (category words, \%)}\par\smallskip
\fitwidth{\begin{tabular}{@{}lrrrr@{}}\toprule
Cue lag $d$ & $r{=}1$ & $r{=}2$ & $r{=}3$ & $r{=}4$\\\midrule
Lag 2 & $49.2\pm0.4$ & $\mathbf{100.0\pm0.0}$ & $100.0\pm0.0$ & $100.0\pm0.0$\\
Lag 3 & $49.2\pm0.4$ & $49.3\pm0.5$ & $\mathbf{100.0\pm0.0}$ & $100.0\pm0.0$\\
Lag 4 & $49.2\pm0.4$ & $49.9\pm0.2$ & $49.9\pm0.2$ & $\mathbf{100.0\pm0.0}$\\\bottomrule
\end{tabular}}
\par\medskip
\textbf{E. Adjacent-path $2\times 2$ restoration (margin)}\par\smallskip
\fitwidth{\begin{tabular}{@{}lrr@{}}\toprule
Adjacent relay & No restoration & Restore anchor ($c_2$)\\\midrule
Open & $0.68\pm1.48$ & $\mathbf{5.60\pm2.23}$\\
Blocked & $-0.10\pm0.79$ & $-0.11\pm0.78$\\\bottomrule
\end{tabular}}
\end{minipage}\hfill
\begin{minipage}[t]{.52\textwidth}\centering
\textbf{D. Full-window edge cuts and $K_{B_1}$ restoration (\%)}\par\smallskip
\fitwidth{\begin{tabular}{@{}lrrr@{}}\toprule
Intervention & Cue at lag 2 & Cue at lag 3 & Cue at lag 4\\\midrule
Clean & $100.0\pm0.0$ & $100.0\pm0.0$ & $100.0\pm0.0$\\
Cut identity edge ($t_i{-}d$) & $49.5\pm0.6$ & $49.3\pm0.5$ & $49.9\pm0.2$\\
Cut word edge ($t_i{-}1$) & $46.1\pm2.3$ & $48.3\pm3.6$ & $47.0\pm1.9$\\
Cut filler edge & $100.0\pm0.0$ & $100.0\pm0.0$ & $100.0\pm0.0$\\
Cut identity + Restore $K_{B_1}$ & $\mathbf{100.0\pm0.0}$ & $\mathbf{100.0\pm0.0}$ & $\mathbf{100.0\pm0.0}$\\
Cut identity + Restore $V_{B_1}$ & $48.6\pm1.6$ & $49.3\pm1.5$ & $50.3\pm1.4$\\\bottomrule
\end{tabular}}
\par\medskip
\textbf{F. Pre-receiver 79M SWA horizon ($R_{\mathrm{local}}=3(W{-}1)$)}\par\smallskip
\fitwidth{\begin{tabular}{@{}lrrrr@{}}\toprule
Window $W$ & $C_1$ (peak) & $\Sigma_3^{\mathrm{blk1}}$ ($\le 3$) & $\Sigma_4^{\mathrm{blk1}}$ ($>3$) & $\Sigma_4^{\mathrm{all}}$\\\midrule
$W{=}2$ ($R{=}3$) & $1.01\pm0.01$ & $0.84\pm0.06$ & $\mathbf{0.00\pm0.00}$ & $0.01\pm0.00$\\
$W{=}4,8,16$ ($R{\ge}9$) & $1.15\text{--}1.37$ & $0.95\text{--}1.20$ & $0.46\text{--}0.53$ & $0.46\text{--}0.53$\\\bottomrule
\end{tabular}}
\end{minipage}
\par\smallskip\parbox{\textwidth}{\footnotesize $^\dagger$With distinct wrong codes per distractor (up to 7 wrong codes), accuracy across 17/29/53/101 tokens is $80.5\pm10.9\%$, $61.2\pm14.2\%$, $39.3\pm12.2\%$, and $26.4\pm13.5\%$. In Panel~F, individual $W{=}4,8,16$ values are $C_1=(1.15,1.37,1.36)$, $\Sigma_3^{\mathrm{blk1}}=(0.95,1.20,1.16)$, and $\Sigma_4^{\mathrm{blk1}}=(0.46,0.50,0.53)$.}
\end{table*}

\subsection{Stage Recruitment, Query Bridge, Corpus Audits, and Relational Transfer}
\label{app:transfer}
\begin{figure}[!t]\centering
\includegraphics[width=\linewidth]{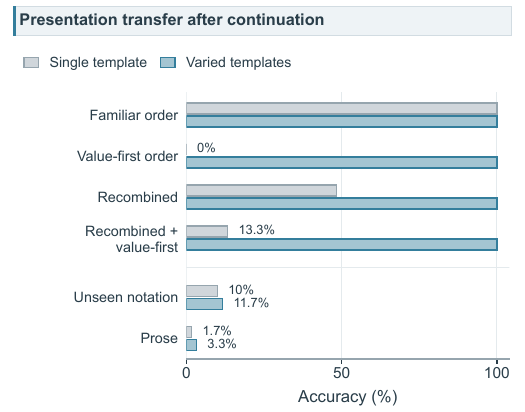}
\caption{\textbf{Template diversity improves transfer within the trained systems.} Matched probes compare single-template and varied-template continuation. The separated bottom group tests unseen notations and prose. Bars show reported aggregate accuracies; corresponding general-language and code perplexities appear in Table~\ref{tab:relational-ppl}F.}
\label{fig:appendix-transfer}
\end{figure}

Table~\ref{tab:recruit} collects the four sequential-retrieval control suites:
\begin{enumerate}[leftmargin=*,nosep]
\item \emph{Recruitment vs.\ task exposure (Table~\ref{tab:recruit}A,B)}: Stagewise-to-serial training preserves Stage-2 Carrying ($5.15\pm 1.57$) and backup recall when Stage~1 is cut ($8.07\pm 2.33$), whereas composite-task exposure improves composite accuracy under both schedules without recruiting Stage~2 under serial training.
\item \emph{Residual relay vs.\ explicit-query positive control (Table~\ref{tab:recruit}C,D)}: In the learned residual-stream suite, ablating $\mathrm{L1}$ disrupts both hops whereas ablating $\mathrm{L3}$ disrupts only the endpoint, and patching only the swapped donor's $\mathrm{L1}$ query state redirects the endpoint prediction ($.912$ vs.\ $.004$ at record positions; Panel~C). In the separate explicit-query positive control (Panel~D), competing intermediate swaps ($b_1\to b_2$) reverse the margin ($+10.61\to -10.59$), unrelated valid intermediates ($b_3$) collapse the margin to zero ($-0.19$), and open access to the second-hop source doubles query-restoration gain ($21.19$ vs.\ $8.82$).
\item \emph{Natural-corpus shortcut audits and 318M continued training (Table~\ref{tab:recruit}E,F)}: Across 800 OpenWebText windows, progressive removal of single-hop and recency shortcuts reduces candidate survival to $0/800$ at \texttt{L6}, and manual inspection confirms that all $6/6$ multi-stage AND-gate survivors are verbatim repeated spans; Python AST alias scans similarly leave $0/42$ conflict-free two-hop instances (Panel~E). In 318M continued training (Fig.~\ref{fig:appendix-transfer}; Table~\ref{tab:recruit}F), template diversity enables held-out symbol recombination and value-first reversal ($1.000$), but does not transfer to unseen notations ($.117$) or prose ($.033$) and increases code perplexity ($34.71\to 38.73$).
\end{enumerate}

\begin{table*}[!t]
\caption{\textbf{Stage recruitment, sequential query-bridge interventions, natural-corpus shortcut audits, and 318M relational transfer.} \textbf{A,B}: Five-seed controlled recruitment and composite-task exposure comparisons. \textbf{C,D}: Learned residual-stream receiver ablations and $\mathrm{L1}\to\mathrm{L3}$ relay patching, alongside explicit-query specificity and second-source blocking controls (5 seeds). \textbf{E}: OpenWebText 7-level difficulty ladder ($n{=}800$) and Python AST alias audit. \textbf{F}: Language/code perplexity after 318M relational continuation. Fig.~\ref{fig:appendix-transfer} shows transfer across presentation types.}
\label{tab:recruit}\label{tab:task-exposure}\label{tab:querycontrols}\label{tab:bridgeinteraction}\label{tab:corpus-ladder}\label{tab:relational-transfer}\label{tab:relational-ppl}
\centering\small\setlength{\tabcolsep}{3pt}
\begin{minipage}[t]{.47\textwidth}\centering
\textbf{A. Stage recruitment (5 seeds)}\par\smallskip
\fitwidth{\begin{tabular}{@{}lrrr@{}}\toprule
Schedule & Stage 1 Carry & Stage 2 Carry & Stage 1 cut margin\\\midrule
Serial & $10.85\pm1.48$ & $0.00\pm0.00$ & $-0.14\pm0.58$\\
Stagewise-to-serial & $7.26\pm2.26$ & $\mathbf{5.15\pm1.57}$ & $\mathbf{8.07\pm2.33}$\\\bottomrule
\end{tabular}}
\par\medskip
\textbf{C. Residual-stream ablations \& $\mathrm{L1}\to\mathrm{L3}$ relay patch}\par\smallskip
\fitwidth{\begin{tabular}{@{}lr@{}}\toprule
Mechanism intervention & Effect / Redirected acc.\\\midrule
Ablate Stage-1 receiver $\mathrm{L1}$ (Hop-1 / Endpoint) & $-.902\ /\ -.455$\\
Ablate Stage-2 receiver $\mathrm{L3}$ (Hop-1 / Endpoint) & $.000\ /\ -.485$\\
Patch swapped donor $\mathrm{L1}$ \textbf{query only} ($c_2$ acc) & $\mathbf{.912}$\\
Patch swapped donor $\mathrm{L1}$ full sequence ($c_2$ acc) & $.912$\\
Patch swapped donor $\mathrm{L1}$ record positions ($c_2$ acc) & $.004$\\
Reverse patch clean $\mathrm{L1}$ query state ($c_1$ acc) & $.997$\\\bottomrule
\end{tabular}}
\end{minipage}\hfill
\begin{minipage}[t]{.51\textwidth}\centering
\textbf{B. Composite-task exposure after recruitment (5 seeds)}\par\smallskip
\fitwidth{\begin{tabular}{@{}llrr@{}}\toprule
Schedule & Exposure & Stage 2 Carry & Composite acc.\\\midrule
Serial & None / $20\%$ & $0.00\ /\ 0.00$ & $.773\pm.045\ /\ \mathbf{.945\pm.039}$\\
Stagewise & None / $20\%$ & $\mathbf{5.15\ /\ 5.04}$ & $.672\pm.063\ /\ \mathbf{.931\pm.047}$\\\bottomrule
\end{tabular}}
\par\medskip
\textbf{D. Explicit-query specificity \& second-source blocking (margin)}\par\smallskip
\fitwidth{\begin{tabular}{@{}lrlrrr@{}}\toprule
Query patch & Margin & Source $2$ & Supply $b_2$ & Supply $b_1$ & Gain\\\midrule
Clean ($b_1$) & $+10.61\pm.85$ & Open & $-10.59$ & $+10.60$ & $\mathbf{21.19\pm1.63}$\\
Swap ($b_1{\to}b_2$) & $-10.59\pm.80$ & Blocked & $-10.82$ & $-2.01$ & $8.82\pm1.33$\\
Unrelated ($b_3$) & $-0.19\pm.44$ & \multicolumn{4}{l}{\footnotesize Off-site ($q{-}8$): $+10.15\pm.57$; Random: $-.35\pm.68$}\\\bottomrule
\end{tabular}}
\end{minipage}
\par\medskip
\begin{minipage}[t]{.52\textwidth}\centering
\textbf{E. Natural-corpus two-hop audits (OWT $n{=}800$ \& Python AST)}\par\smallskip
\fitwidth{\begin{tabular}{@{}llrr@{}}\toprule
Level / Filter & Cumulative requirement / Rule & Left & Share / Audit finding\\\midrule
\texttt{L0}$\to$\texttt{L1} & Valid $A{\to}B{\to}C$; hop 2 visible & $800{\to}613$ & $100\%\to 76.6\%$\\
\texttt{L2}$\to$\texttt{L3} & Bridge $\le 2$; latest $B$ is target & $104{\to}76$ & $13.0\%\to 9.5\%$\\
\texttt{L4}$\to$\texttt{L5} & Unique $C$ ($\mathrm{L4}$); unique $A$ succ.\ ($\mathrm{L5}$) & $9{\to}3$ & $1.1\%\to 0.4\%$\\
\texttt{L6} & Nonlocal $B,C$ ($>8$ tok); no dup.\ span & \textbf{0} & $\mathbf{0.0\%}$ (\texttt{L5} all $\mathrm{D0}$)\\\midrule
OWT funnel & \texttt{hop1\_maj} ($46$) $\to$ \texttt{hop2\_maj} ($9$) $\to$ clean & \textbf{6} & \textbf{6/6 repeated spans}\\
Python AST & $42$ aliases $\to$ \texttt{no\_copy} ($13$) $\to$ conflict & \textbf{0} & $29/42$ copy; $59.5\%$ \texttt{.parent}\\\bottomrule
\end{tabular}}
\end{minipage}\hfill
\begin{minipage}[t]{.46\textwidth}\centering
\textbf{F. 318M continuation perplexity}\par\smallskip
\fitwidth{\begin{tabular}{@{}lrr@{}}\toprule
Continuation data & General & Code\\\midrule
Language only & $27.08$ & $34.71$\\
Single template & $27.27$ & $35.61$\\
Varied templates & $27.29$ & $38.73$\\\bottomrule
\end{tabular}}
\end{minipage}
\end{table*}

\section{Locality and Retrieval Learning}
\label{app:theory}
The theoretical analysis uses the same source-address ($K_j=A_r(X)_j$) and sequential-retrieval interfaces as the experiments. Propositions~\ref{prop:locality}--\ref{prop:memorization} provide the formal statements and proofs referenced in Sections~\ref{sec:interface}--\ref{sec:sequential}. Table~\ref{tab:context-access}C,D give the empirical counterpart to the source-horizon construction.

\subsection{Address Construction and Relays}
\begin{proposition}[Locality and adjacent composition]
\label{prop:locality}
Consider $L$ local causal layers, each using offsets $0,\ldots,W-1$, with no cross-position recurrent state. The representation at position $t$ depends only on $x_{t-L(W-1):t}$. Hence a source key built from this stack is invariant to a cue swap outside that interval. If query and other input routes are fixed, key-based matching is invariant too.

Let $S$ be the causal one-position shift, with $\|S\|\le1$. If local operators satisfy $\|\widehat S_i\|\le1$ and $\|\widehat S_i-S\|\le\epsilon_i$, then
\begin{equation}
 \|\widehat S_r\cdots\widehat S_1-S^r\|\le\sum_{i=1}^{r}\epsilon_i.
\end{equation}
\end{proposition}
\begin{proof}
After one layer the accessible offsets are $0,\ldots,W-1$. Each additional layer adds at most $W-1$ preceding positions, proving the radius $R_{\mathrm{local}}\le L(W-1)$ by induction. A change outside that set cannot change the representation or any deterministic source-key projection. Fixed queries therefore yield identical key-based matching scores. Let $S$ denote the causal one-position shift, $(SX)_t=X_{t-1}$. For approximate shifts, expand the product difference as
\[
 \sum_{i=1}^{r}\widehat S_r\cdots\widehat S_{i+1}
 (\widehat S_i-S)S^{i-1}.
\]
The triangle inequality and contraction bounds give the stated sum.
\end{proof}

\begin{proposition}[Source-horizon indistinguishability and composite address sufficiency]
\label{prop:indistinguishability}
Consider paired prompts $x,x'$ with two category-compatible records $a,b$ and distinct codes $y_a\ne y_b$. The pair swaps only their identity cues $c_a=x_{t_a-d}$ and $c_b=x_{t_b-d}$, with $d\ge2$, while leaving cue words $w_i=x_{t_i-1}$, codes, and the query tokens fixed. The query selects identity $c_a$ in their shared category, so its correct code changes from $y_a$ to $y_b$. A single local writer precedes a global reader that attends only to code positions; later blocks are positionwise. Assume the query representation is invariant to the swap and no other path carries the swapped identities to the output. Then:
\begin{enumerate}[leftmargin=*,nosep]
\item Necessity: If every reader-visible source representation excludes both swapped identity positions, any deterministic model has average accuracy at most $1/2$ on the pair. A radius $r<d$ gives this exclusion when record spacing also keeps other records' identities outside every source window. An edge cut gives the same result when it removes every path from the swapped identities to the reader-visible sources.
\item Sufficiency: Suppose four records form the Cartesian product of two identities and two categories, with one code per combination, and $r\ge d$. With orthonormal one-hot features, set $K_i=e_{c_i}\oplus e_{g(w_i)}$, $Q_q=\beta(e_{c_q}\oplus e_{g(w_q)})$, and $V_i=e_{y_i}$. The unique matching record receives attention $\alpha_*=e^{2\beta}/(e^{2\beta}+2e^\beta+1)\to1$ as $\beta\to\infty$. This holds for both identical cue words and different words in the same category.
\end{enumerate}
\end{proposition}
\begin{proof}
For Part~(1), the access assumptions imply identical keys, values, and query representations in $x,x'$. The reader and subsequent positionwise computation therefore produce the same prediction. Since the targets differ, at most one member of each pair is correct. In our controlled construction, code positions are six tokens apart, identity offsets are at most four, and the query's local window contains no swapped identity; thus $r<d$ excludes the changed information from all reader inputs. The single-layer writer also makes its identity-edge mask a complete cut of that route. For Part~(2), $Q_q^\top K_i/\beta$ equals two on the unique target, one on each record matching only identity or category, and zero on the remaining record. Softmax gives the stated weight, which converges to one.
\end{proof}

\subsection{Shared Parameters and Inactive Multi-Gate Paths}
\label{app:gates}
Equation~\eqref{eq:conflict} follows by Taylor expansion of $\mathcal L_1(w-\eta(\pi_0g_0+\pi_1g_1))$, separating the self-gradient term from interference through shared parameters. For a cross-layer path $\theta_1\theta_2\theta_3$ that shares residual target $y>0$ with an alternative intra-layer state readout $u$, consider the coupled regularized surrogate
\begin{align}
 \mathcal J(u,\theta)={}&\tfrac12(y - u - \theta_1\theta_2\theta_3)^2
 +\tfrac\mu2 u^2\nonumber\\
 &+\tfrac\lambda2\sum_{i=1}^3\theta_i^2,\qquad y,\mu,\lambda>0.
\end{align}
Minimizing with respect to the alternative readout $u$ yields $u^*(\theta)=\frac{y-\theta_1\theta_2\theta_3}{1+\mu}$, which absorbs a $\frac{1}{1+\mu}$ fraction of the residual and reduces the effective target driving the cross-layer path to $\rho_{\mathrm{eff}}=\frac{\mu}{1+\mu}y < y$:
\begin{equation}
 \mathcal J(u^*(\theta),\theta)=\frac{\mu}{2(1+\mu)}\Big(y-\theta_1\theta_2\theta_3\Big)^2+\frac{\lambda}{2}\sum_{i=1}^3\theta_i^2.
\end{equation}
At the inactive cross-layer state $(u_*,0,0,0)$ with $u_*=\frac{y}{1+\mu}$, the gradient with respect to $\theta$ is zero and the Hessian is $\operatorname{diag}(\lambda,\lambda,\lambda)\succ 0$, so the origin is a strict local minimum. Holding $\theta_1=\Theta_1$ changes the remaining two-gate Hessian eigenvalues to $\lambda\pm\frac{\mu}{1+\mu}y|\Theta_1|$; a negative curvature escape direction appears only once $|\Theta_1|>\frac{\lambda(1+\mu)}{\mu y}$. Thus, stronger residual absorption by the alternative state route (smaller $\mu$) raises the threshold required for the cross-layer path to escape the origin.

\subsection{Task Demand and New Computational Directions}
\begin{proposition}[Incremental demand and a gated path]
\label{prop:demand}
Let $\mathcal H_{k-1}\subseteq\mathcal H_k\subset L^2(D)$ be finite-dimensional linear prediction spaces under risk $R(f)=\frac12\mathbb E[(Y-f(X))^2]$. Let $f_k^*$ be the least-squares projection and $d_k=f_k^*-f_{k-1}^*$. Then
\begin{equation}
 \Delta_k=R(f_{k-1}^*)-R(f_k^*)=\tfrac12\|d_k\|_{L^2(D)}^2.
\end{equation}
The largest absolute functional derivative at $f_{k-1}^*$ over directions in $\mathcal V_k=\mathcal H_k\cap\mathcal H_{k-1}^{\perp}$ of norm at most one is $\sqrt{2\Delta_k}$. For positive demand, a unit maximizing direction and amplitude $a=\prod_e\theta_e$ give the data gradient in Eq.~\eqref{eq:gates}.
\end{proposition}
\begin{proof}
The projection residual $Y-f_k^*$ is orthogonal to $\mathcal H_k$, while $d_k\in\mathcal V_k$. Pythagoras gives the risk difference. For unit $\psi\in\mathcal V_k$,
\[
 \left.\frac{d}{da}R(f_{k-1}^*+a\psi)\right|_{a=0}
 =-\langle d_k,\psi\rangle.
\]
Cauchy--Schwarz bounds its magnitude by $\|d_k\|=\sqrt{2\Delta_k}$, attained by $\psi=d_k/\|d_k\|$ when demand is positive. Along this direction the risk is a constant plus $\frac12(a-\sqrt{2\Delta_k})^2$; applying the chain rule to $a=\prod_e\theta_e$ gives Eq.~\eqref{eq:gates}.
\end{proof}

\subsection{Residual Risk and Memorization}
\begin{proposition}[Unresolved risk gap and fixed-instance lookup attenuation]
\label{prop:memorization}
Suppose a model's active reusable path features lie in $\mathcal H_K$, whereas target distribution $D$ requires features up to $\mathcal H_H$ ($H>K$). Then:
\begin{enumerate}[leftmargin=*,nosep]
\item The unresolved risk gap is $R_D^*(K)-R_D^*(H)=\sum_{k=K+1}^H\Delta_k(D)$, and adding parallel modules whose features remain in $\mathcal H_K$ leaves this gap unchanged.
\item In a model augmented with an instance-indexed lookup vector $m\in\mathbb R^N$ bound to $N$ fixed training instances ($m=0$ on unseen or role-relabeled instances) under
\begin{equation}
 \mathcal J(\theta,m)=\frac{1}{2N}\|y-f_\theta-m\|_2^2+\frac{\lambda}{2}\|\theta\|_2^2+\frac{\mu}{2N}\|m\|_2^2,
\end{equation}
where $\mu>0$, the optimal lookup $m^*=(y-f_\theta)/(1+\mu)$ reduces the structural objective to
\begin{equation}\label{eq:prop3-att}
 \mathcal J(\theta,m^*(\theta))=\frac{\mu}{1+\mu}\cdot\frac{1}{2N}\|y-f_\theta\|_2^2+\frac{\lambda}{2}\|\theta\|_2^2,
\end{equation}
absorbing a $1/(1+\mu)$ fraction of the training residual and attenuating the data-loss gradient with respect to $\theta$ by $\frac{\mu}{1+\mu}<1$. The parameter-regularization gradient is unchanged.
\end{enumerate}
\end{proposition}
\begin{proof}
Part (1) follows by telescoping $\Delta_k(D)=R_D^*(k-1)-R_D^*(k)$ from $k=K+1$ to $H$ and noting that $\operatorname{span}(\mathcal H_K\cup\mathcal H_K)=\mathcal H_K$. For Part (2), setting $\nabla_m\mathcal J = \frac{1}{N}(m-(y-f_\theta))+\frac{\mu}{N}m = 0$ yields $m^*=(y-f_\theta)/(1+\mu)$ and residual $y-f_\theta-m^*=\frac{\mu}{1+\mu}(y-f_\theta)$. Substituting $m^*$ into $\mathcal J(\theta,m)$ gives $\frac{1}{2N}(\frac{\mu^2}{(1+\mu)^2}+\frac{\mu}{(1+\mu)^2})\|y-f_\theta\|_2^2=\frac{\mu}{1+\mu}\frac{1}{2N}\|y-f_\theta\|_2^2$, which scales the data gradient with respect to $\theta$ by $\frac{\mu}{1+\mu}<1$.
\end{proof}

\section{Architectural Implications}
\label{app:architecture}
The tested interfaces suggest the following architectural hypotheses and corresponding interventions.

\paragraph{Local and recurrent layers}
Sliding-window layers relay context within a finite horizon; recurrent state also carries information across that horizon. Source-window lesions and source-key restoration distinguish address preparation from direct state readout.

\paragraph{Sparse candidate access}
Sparse attention~\citep{yuan2025nsa} must include the target before selecting it. Measuring candidate inclusion and conditional target preference separately tests whether failure occurs during sparse selection or address matching.

\paragraph{Conditional lookup features}
Engram-style $n$-gram features~\citep{cheng2026engram} may supply retrieval addresses. Swapping lookup features, restoring receiver keys, and blocking native Carrying would distinguish replacement from complementary input; the lookup's token coverage and output position matter.

\paragraph{Recurrence and intermediate tokens}
Loops and generated intermediates can implement $q_{i+1}=\mathcal B(y_i)$~\citep{dehghani2019universal,saunshi2025looped,wei2022cot}. Intermediate swaps followed by second-source interventions test whether additional execution steps form causally sequential retrievals.

\end{document}